\documentclass[journal]{IEEEtran}
\ifCLASSINFOpdf
\else
\fi

\usepackage{bm}
\usepackage{times}
\usepackage{epsfig}
\usepackage{graphicx}
\usepackage{amsmath}
\usepackage{amssymb}
\usepackage{subfigure}
\usepackage{caption}
\usepackage{multirow}
\usepackage{threeparttable}
\usepackage{arydshln}
\usepackage{color}
\usepackage{cite}
\usepackage{url}
\usepackage[colorlinks,urlcolor=black,linkcolor=black,anchorcolor=black,citecolor=black]{hyperref}
\begin{document}
%
% paper title
% Titles are generally capitalized except for words such as a, an, and, as,
% at, but, by, for, in, nor, of, on, or, the, to and up, which are usually
% not capitalized unless they are the first or last word of the title.
% Linebreaks \\ can be used within to get better formatting as desired.
% Do not put math or special symbols in the title.
%\title{Noise-Robust Image Classification by Suppressing Aliasing Effect Using Wavelet Transformed CNN}
\title{WaveCNet: Wavelet Integrated CNNs to Suppress Aliasing Effect for Noise-Robust Image Classification}
%
%
% author names and IEEE memberships
% note positions of commas and nonbreaking spaces ( ~ ) LaTeX will not break
% a structure at a ~ so this keeps an author's name from being broken across
% two lines.
% use \thanks{} to gain access to the first footnote area
% a separate \thanks must be used for each paragraph as LaTeX2e's \thanks
% was not built to handle multiple paragraphs
%

\author{Qiufu~Li,~~
        Linlin~Shen\textsuperscript{*},~~
        Sheng~Guo,~~
        Zhihui~Lai% <-this % stops a space
% <-this % stops a space
\thanks{The work is supported by the Natural Science Foundation of China under grants no. 62006156, 91959108 and U1713214,
and the Science and Technology Project of Guangdong Province under grant no. 2018A050501014.
Corresponding author: Linlin Shen.}
\thanks{Q. Li, L. Shen, and Z. Lai are with the Computer Vision Institute,
College of Computer Science and Software Engineering, Shenzhen University, Shenzhen 518060, China,
Shenzhen Institute of Artificial Intelligence and Robotics for Society, Shenzhen 518060, China,
and Guangdong Key Laboratory of Intelligent Information Processing, Shenzhen University, Shenzhen 518060, China
(e-mail: qiufu\_li\_1988@163.com; llshen@szu.edu.cn; lai\_zhi\_hui@163.com).
S. Guo is with MyBank, Ant Group, Hangzhou 310012, China (e-mail: guosheng.guosheng@alibaba-inc.com).}% <-this % stops a space
}

% note the % following the last \IEEEmembership and also \thanks -
% these prevent an unwanted space from occurring between the last author name
% and the end of the author line. i.e., if you had this:
%
% \author{....lastname \thanks{...} \thanks{...} }
%                     ^------------^------------^----Do not want these spaces!
%
% a space would be appended to the last name and could cause every name on that
% line to be shifted left slightly. This is one of those "LaTeX things". For
% instance, "\textbf{A} \textbf{B}" will typeset as "A B" not "AB". To get
% "AB" then you have to do: "\textbf{A}\textbf{B}"
% \thanks is no different in this regard, so shield the last } of each \thanks
% that ends a line with a % and do not let a space in before the next \thanks.
% Spaces after \IEEEmembership other than the last one are OK (and needed) as
% you are supposed to have spaces between the names. For what it is worth,
% this is a minor point as most people would not even notice if the said evil
% space somehow managed to creep in.

% The paper headers
\markboth{IEEE Transactions on Image Processing}%
{Li \MakeLowercase{\textit{et al.}}: WaveCNet: Wavelet Integrated CNNs to Suppress Aliasing Effect for Noise-Robust Image Classification}
% The only time the second header will appear is for the odd numbered pages
% after the title page when using the twoside option.
%
% *** Note that you probably will NOT want to include the author's ***
% *** name in the headers of peer review papers.                   ***
% You can use \ifCLASSOPTIONpeerreview for conditional compilation here if
% you desire.

% If you want to put a publisher's ID mark on the page you can do it like
% this:
%\IEEEpubid{0000--0000/00\$00.00~\copyright~2015 IEEE}
% Remember, if you use this you must call \IEEEpubidadjcol in the second
% column for its text to clear the IEEEpubid mark.

% use for special paper notices
%\IEEEspecialpapernotice{(Invited Paper)}

% make the title area
\maketitle

% As a general rule, do not put math, special symbols or citations
% in the abstract or keywords.
\begin{abstract}
Though widely used in image classification, convolutional neural networks (CNNs) are prone to noise interruptions,
i.e. the CNN output can be drastically changed by small image noise.
To improve the noise robustness, we try to integrate CNNs with wavelet
by replacing the common down-sampling (max-pooling, strided-convolution, and average pooling) with discrete wavelet transform (DWT).
We firstly propose general DWT and inverse DWT (IDWT) layers applicable to various orthogonal and biorthogonal discrete wavelets like Haar, Daubechies, and Cohen, etc.,
and then design wavelet integrated CNNs (WaveCNets) by integrating DWT into the commonly used CNNs (VGG, ResNets, and DenseNet).
During the down-sampling, WaveCNets apply DWT to decompose the feature maps into the low-frequency and high-frequency components.
Containing the main information including the basic object structures, the low-frequency component is transmitted into the following layers to generate robust high-level features.
The high-frequency components are dropped to remove most of the data noises.
The experimental results show that
%wavelet accelerates the CNN training, and
WaveCNets achieve higher accuracy on ImageNet than various vanilla CNNs.
We have also tested the performance of WaveCNets on the noisy version of ImageNet, ImageNet-C and six adversarial attacks,
the results suggest that the proposed DWT/IDWT layers could provide better noise-robustness and adversarial robustness.
{\color{black}When applying WaveCNets as backbones, the performance of object detectors (i.e., faster R-CNN and RetinaNet) on COCO detection dataset are consistently improved.}
We believe that suppression of aliasing effect, i.e. separation of low frequency and high frequency information, is the main advantages of our approach.
The code of our DWT/IDWT layer and different WaveCNets are available at {\href{https://github.com/LiQiufu/WaveCNet}{https://github.com/CVI-SZU/WaveCNet}}.
\end{abstract}

% Note that keywords are not normally used for peerreview papers.

% For peer review papers, you can put extra information on the cover
% page as needed:
% \ifCLASSOPTIONpeerreview
% \begin{center} \bfseries EDICS Category: 3-BBND \end{center}
% \fi
%
% For peerreview papers, this IEEEtran command inserts a page break and
% creates the second title. It will be ignored for other modes.
\IEEEpeerreviewmaketitle

\section{Introduction}
% The very first letter is a 2 line initial drop letter followed
% by the rest of the first word in caps.
%
% form to use if the first word consists of a single letter:
% \IEEEPARstart{A}{demo} file is ....
%
% form to use if you need the single drop letter followed by
% normal text (unknown if ever used by the IEEE):
% \IEEEPARstart{A}{}demo file is ....
%
% Some journals put the first two words in caps:
% \IEEEPARstart{T}{his demo} file is ....
%
% Here we have the typical use of a "T" for an initial drop letter
% and "HIS" in caps to complete the first word.
\IEEEPARstart{S}{mall} noise, including the common spatial noise \cite{hendrycks2019benchmarking}
and the specially designed adversarial noise \cite{goodfellow2014explaining,Kurakin2017adversarial,Tramer2018ensemble,Carlini2016towards,Madry2016towards},
can drastically change the final predication of well trained convolutional neuronal network (CNN) for image classification.
The recent studies \cite{xie2019feature, geirhos2018imagenet} show that the noise may be enlarged as the image data flows through the deep networks.
These phenomena illustrate the weak noise-robustness of CNNs.

\begin{figure}[!tbp]
	\centering
	\includegraphics*[scale=0.725, viewport=20 62 366 380]{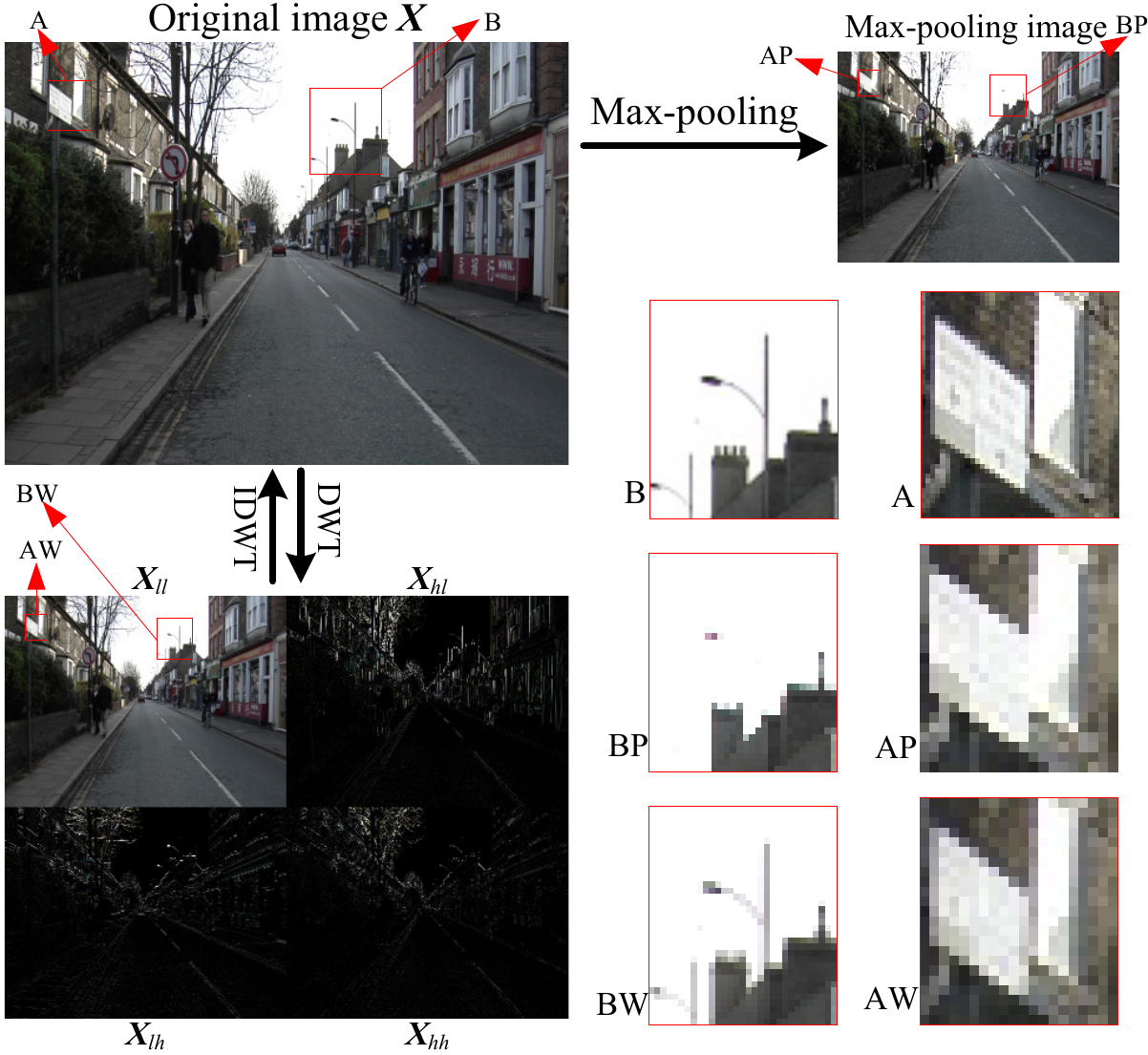}
	\caption{Comparison of max-pooling and wavelet transforms.
		Max-pooling is a commonly used down-sampling operation in the deep networks,
		which could easily breaks the basic object structures.
		Discrete Wavelet Transform (DWT) decomposes an image $\textbf{\emph{X}}$ into its low-frequency component $\textbf{\emph{X}}_{ll}$
		and high-frequency components $\textbf{\emph{X}}_{lh}, \textbf{\emph{X}}_{hl}, \textbf{\emph{X}}_{hh}$.
		While $\textbf{\emph{X}}_{lh}, \textbf{\emph{X}}_{hl}, \textbf{\emph{X}}_{hh}$ represent image details including most of the noise,
		$\textbf{\emph{X}}_{ll}$ is a low resolution version of the image, where the basic object structures are represented.
		In the figures, the window boundary in area A (AP) and the poles in area B (BP) are broken by max-pooling,
		while the principal features of these objects are kept in the DWT output (AW and BW).
		%In addition, the background color of area AP somewhat differs from that of area A and AW,
		%which may be resulted from the noise accumulation during max-pooling.
		%Best viewed in color.
	}\label{fig_DWT}
\end{figure}
The weak noise-robustness of CNNs is closely related to the down-sampling.
The commonly used down-sampling operations in deep networks,
such as max-pooling, average-pooling, and strided-convolution,
usually ignore the classic sampling theorem \cite{nyquist1928certain},
which result in aliasing among the data components in different frequency intervals.
While the noise of data is mostly high-frequency components,
the low-frequency component contains the main information, such as the basic object structures.
Therefore, the aliasing introduces residual noise in the down-sampled data and breaks the basic structures,
which degrades the accuracy and noise-robustness of CNNs.
Fig. \ref{fig_DWT} presents a max-pooling example.

In signal processing, to avoid the aliasing, the data is usually decomposed into different frequency intervals
using time-frequency analysis tools, such as wavelet \cite{mallat1989theory,daubechies1992ten}.
Discrete wavelet transform (DWT), consisting of filtering and down-sampling, decomposes a 2D data into four low-frequency and high-frequency components, as Fig. \ref{fig_DWT} shows.
%The basic object structures and most of noise are represented in the low-frequency and high-frequency components, respectively.
DWT separates the main information and details of the data,
while inverse DWT (IDWT), consisting of up-sampling and filtering, reconstructs the original data using the DWT output.

In this paper, to suppress the aliasing effect in CNNs for noise-robust image classification,
we integrate the commonly used CNN architectures with discrete wavelet transform.
We firstly analyze the data forward and backward propagation in the wavelet transforms (DWT and IDWT),
and rewrite them as general network layers in PyTorch \cite{paszke2017automatic}.
Then, we design wavelet integrated convolutional networks (WaveCNets).
In WaveCNets, during the down-sampling, the feature maps are decomposed by DWT into the low-frequency and high-frequency components.
While the low-frequency component is transmitted to the following layers for robust high-level features,
the high-frequency components are dropped to resist the noise propagation.
WaveCNets are evaluated on ImageNet \cite{deng2009imagenet} and COCO \cite{COCO_2014_ECCV}
in terms of classification accuracy, noise-robustness, adversarial robustness, and detection precision,
when various discrete wavelets and CNN architectures are used.
In summary:
\begin{enumerate}
\setlength{\topsep}{0ex}
\setlength{\itemsep}{0ex}
\item We design general DWT/IDWT layer applicable to various discrete wavelets, which could be used to design end-to-end wavelet integrated deep networks.
\item We propose WaveCNets by transforming the feature maps using DWT during down-sampling
	  to suppress aliasing effect for noise-robust image classification.
\item {\color{black}Our WaveCNets are evaluated by both classification and detection tasks.
      The results on ImageNet classification show that our approach can achieve higher accuracy, better noise-robustness, and increased adversarial robustness.
      The detection results on COCO benchmark suggest that faster R-CNN \cite{faster_rcnn} and RetinaNet \cite{Lin_2017_RetinaNet}, using WaveCNets as the backbones,
      achieve better performance on detecting small, medium, and large objects.}
%\item WaveCNets are evaluated on ImageNet,
%	  and achieve higher accuracy, better noise-robustness, and increased adversarial robustness.
%\item {\color{black}Applying WaveCNets as backbones instead of the vanilla CNNs,
%      the image detection performance of faster R-CNN and RetinaNet are consistently improved. }
\end{enumerate}

An earlier version of this paper \cite{qiufu_2020_CVPR}
has been accepted by the 2020 IEEE Conference on Computer Vision and Pattern Recognition (CVPR2020).
We have extensively extended the contents by presenting the introduction of wavelet theory,
exploring the details of 2D DWT/IDWT layers and giving the details of backward propagation of both 1D and 2D DWT/IDWT layers.
For experiments, we tested the performance of using wavelet denoising as a preprocessing step and the comparison shows that our DWT/IDWT layers achieves significantly better noise-robustness.
We have also tested the performance of our WaveCNets against adversarial attacks.
As expected, the results suggest that our DWT/IDWT layers can reasonably improve the robustness of different CNNs against adversarial samples.
{\color{black}In addition, we also evaluate the performance of WaveCNets based backbones for object detection task using COCO dataset.
%At last, the image detection performances of faster R-CNN and RetinaNet, taking WaveCNets as their backbones,
%are evaluated on COCO benchmark.
The results show that wavelet could consistently improve the detection performances on small, medium, and large objects.}
%\begin{enumerate}
%\setlength{\topsep}{0ex}
%\setlength{\itemsep}{0ex}
%\item We present the basic wavelet theory in Sec. \ref{subsec_wavelet}.
%\item Sec. \ref{subsec_wavelet_transform_layers} shows the complete analysis for 2D DWT/IDWT layers.
%\item The calculation amount of 2D DWT/IDWT in WaveCNets is analyzed in Sec. \ref{subsec_computation_analysis}.
%\item We illustrate the application of the general wavelet denoising block in the improvement of CNN noise-robustness in Sec. \ref{subsec_wavelet_denoising_block}.
%\item We validate the increased adversarial robustness of WaveCNets in Sec. \ref{subsec_adversarial_robustness}.
%\end{enumerate}

\section{Related works}
\subsection{Noise-robustness}
The recent studies show that ImageNet-trained CNNs prefer to extract features from object textures
sensitive to noise \cite{brendel2019approximating, geirhos2018imagenet},
and the noise could be enlarged as the image data flows through layers in the CNNs \cite{liao2018defense,xie2019feature},
leading to the final wrong predictions.
These works illustrate the weak noise-robustness of modern CNNs.
When the input image is corrupted by noise, the output of CNN can be significantly changed,
regardless of whether the noise is easily perceived by human or not
\cite{hendrycks2019benchmarking,goodfellow2014explaining,Kurakin2017adversarial,Tramer2018ensemble,Carlini2016towards,Madry2016towards}.
The common spatial noise, such as Gaussian noise, shot noise, and impulse noise,
can noticeably degrades the image quality, and decrease the classification accuracy of CNNs \cite{hendrycks2019benchmarking}.
Adversarial noise, produced by specially designed algorithms \cite{goodfellow2014explaining,Kurakin2017adversarial,Tramer2018ensemble,Carlini2016towards,Madry2016towards},
can successfully attack the well trained CNNs,
although the noise is not perceived by human visual system.
A benchmark evaluating CNN performance on noisy images is proposed in \cite{hendrycks2019benchmarking}.
Our WaveCNets will be evaluated using this benchmark.

%When the input image is changed, the output of CNN can be significantly different,
%regardless of whether the change can be easily perceived by human or not
%\cite{goodfellow2014explaining, geirhos2018imagenet, liao2018defense, xie2019feature}.
%While the changes may result from various factors,
%such as shift \cite{zhang2019making,mairal2014convolutional}, rotation \cite{bruna2013invariant_scatnet},
%noise \cite{xie2019feature}, blur \cite{hendrycks2019benchmarking}, manual attack \cite{goodfellow2014explaining}, etc.,
%we mainly focus on the robustness of CNNs to the common noise.
The conventional data augmentation by adding noise to the training images could increase the CNN performance on noisy images \cite{xie2019feature}.
Stylized ImageNet \cite{geirhos2018imagenet} is proposed via stylizing ImageNet images with style transfer
to train the CNNs to extract more robust features from object structures.
The augmentation of training data could noticeably increase the training time or decrease the accuracy on the normal clean images.
The recent studies propose CNN architectures integrating filtering block to denoise feature maps during network inference.
In \cite{liao2018defense}, the authors design a high-level representation guided denoiser
to filter the contaminated images before inputting them into CNN,
which complicates the whole deep network architecture.
A spatial filtering block is presented in \cite{xie2019feature} to denoise the CNN feature maps
and suppress the noise effect on the CNN prediction.
However, the filtering block filter the feature maps using Gaussian filtering, mean filtering, or median filtering,
which do denoising in the whole frequency domain and easily break basic object structure represented by the low-frequency component.
Therefore, this denoising block requires a residual structure for the CNN to converge.

% in the deep networks.

\subsection{Down-sampling}
The weak noise-robustness of CNNs is closely related to down-sampling in the networks.
The down-sampling operations, such as max-pooling, average-pooling, and strided-convolution,
are introduced into deep networks for local connectivity and weight sharing.
These operations could easily erase or dilute image details \cite{yu2014mixed,zeiler2013stochastic}, and then break the basic object structures.
While mixed-pooling \cite{yu2014mixed} and stochastic pooling \cite{zeiler2013stochastic} are proposed to address the drawbacks,
the max-pooling, average-pooling, and strided-convolution are still the most widely applied down-sampling operations in CNNs
\cite{he2016deep,huang2017densely,sandler2018mobilenetv2,simonyan2014very}.

The above down-sampling operations usually ignore the Nyquist's sampling theorem \cite{nyquist1928certain,azulay2018deep,zhang2019making},
which result in aliasing among the data components, break basic object structures and accumulate random noise during CNN inference.
Fig. \ref{fig_DWT} presents a max-pooling example.
Low-pass filtering is integrated with the down-sampling in anti-aliased CNNs \cite{zhang2019making}, to increase the shift-invariance of CNNs.
The author is surprised at the increased classification accuracy and better noise-robustness.
Our WaveCNets and anti-aliased CNNs are significantly different in two aspects:
(1) While anti-aliased CNNs are designed to increase the CNN shift-invariance,
the WaveCNets are conceived on the idea of suppressing aliasing effect using wavelet transform to increase the noise-robustness.
(2) The low-pass filters used in anti-aliased CNNs are empirically designed based on the row vectors of Pascal's triangle,
which is ad hoc and no theoretical justifications are given.
No up-sampling operation, i.e., reconstruction, of the low-pass filter is available.
Therefore, the anti-aliased U-Net \cite{zhang2019making}, the encoder-decoder version of anti-aliased deep network,
has to apply the same filtering after normal up-sampling to achieve the anti-aliasing effect.
In comparison, our WaveCNets are justified by the well defined wavelet theory \cite{daubechies1992ten,mallat1989theory}.
Both the usual down-sampling and up-sampling operations can be replaced by DWT and IDWT \cite{qiufu_2020_CVPR,Qiufu2020WaveSNet}, respectively.

In deep networks for image-to-image translation tasks, the up-sampling operations,
such as max-unpooling in SegNet \cite{badrinarayanan2017segnet},
deconvolution in U-Net \cite{ronneberger2015u_net},
and bilinear interpolation in DeepLab \cite{chen2014semantic_deeplabv1,chen2018encoder_deeplabv3+}
are widely applied to upgrade the feature map resolution.
However, these up-sampling operations can not precisely recover the original data,
due to the absence of the strict mathematical terms.
They do not perform well in the restoration of image details,
while the proposed DWT/IDWT layer could relieve this drawback \cite{Qiufu2020WaveSNet}.

\subsection{Wavelets in deep learning}
Wavelet \cite{daubechies1992ten,mallat1989theory} has wide applications in signal processing, pattern recognition, etc.,
due to its superior performance in time-frequency analysis.
Discrete wavelet transform (DWT) decomposes a data into various components,
and separates the main information and details of the data.
The original data could be reconstructed by inverse DWT (IDWT) using the DWT output.
In signal processing, DWT is a useful tool for anti-aliasing,
and we mainly explore in this paper its application in suppressing the aliasing effect in CNNs for noise-robust image classification.

When Mallat et al explore the optimal deep network from mathematical and algorithmic perspective,
they present ScatNet \cite{bruna2013invariant_scatnet} by cascading wavelet transform with average-pooling and nonlinear modulus operation.
ScatNet preserves the image detail information and extracts a translation invariant feature robust to deformations.
Compared with the CNNs of same period, ScatNet achieves better performance on the texture discrimination and handwritten digit recognition tasks.
In \cite{Thomas2018Mathematical}, Wiatowski et al extend the idea of ScatNet to semi-discrete frames and general nonlinear operations (ReLU, Sigmoid, etc.).
%ScatNet \cite{bruna2013invariant_scatnet} cascades wavelet transform with nonlinear modulus and average-pooling,
%to extract a translation invariant feature robust to deformations and preserve high-frequency information for image classification.
%The authors introduce ScatNet when they explore from mathematical and algorithmic perspective to design the optimal deep network.
However, ScatNet is essentially a hand-designed feature extractor without learnable parameters.
Due to the strict mathematical terms,
ScatNet can not be easily transferred to image-to-image tasks, such as image segmentation.

In the early studies of wavelet integrated neural networks,
the researchers implement wavelet transform using parameterized one-layer networks,
and search the optimal wavelet in parameter domain for function approximation \cite{zhang1992wavelet},
signal representation \cite{szu1992neural}.
The recent work \cite{de2019multi} applies this method with deeper network for image classification.
However, this wavelet parameterized deep network is difficult to train because of the significantly increased computation \cite{de2019multi}.
%Recently, this method is utilized with deeper network for image classification,
%but the network is difficult to train because of the significant amount of computational cost \cite{de2019multi}.

In current deep learning, while wavelet transform is commonly applied as image preprocessing or postprocessing
\cite{huang2017wavelet, liu2018attribute, savareh2019wavelet, yuan2019waveletfcnn},
it is also used as down-sampling or up-sampling operations to design deep netwotks
\cite{liu2018multi,Williams2018Wavelet,duan2017sar,yoo2019photorealistic}.
Multi-level wavelet CNN (MWCNN) \cite{liu2018multi} is a wavelet integrated encoder-decoder for image restoration,
which implements wavelet package transform (WPT) by processing the concatenation of various components of the input data in a unified way.
However, the details represented by the high-frequency components may not be perceived by MWCNN,
because the data amplitude of the high-frequency components is much smaller than that of low-frequency one.
In \cite{duan2017sar}, the authors apply dual-tree complex wavelet transform (DT-CWT)
and design convolutional-wavelet neural network (CWNN),
to extract robust features from SAR images.
CWNN adopts DT-CWT to suppress the noise and keep the object structures in the SAR images,
which contains only two convolutional layers.
DT-CWT is redundant,
and the average value of the two low-frequency components is taken as the down-sampling output of CWNN.
In \cite{Williams2018Wavelet}, the authors propose wavelet pooling layer using a two-level DWT and a one-level IDWT,
while the back-propagation is implemented using one-level DWT and a two-level IDWT, which does not follow the mathematical principle of gradient.
The author design wavelet integrated networks and evaluate them on various datasets
(MNIST \cite{lecun1998gradient}, CIFAR-10 \cite{krizhevsky2009learning}, SHVN \cite{netzer2011reading},
and KDEF \cite{lundqvist1998karolinska}).
However, these wavelet integrated networks consist of only four or five layers,
and they are not systematically studied on the standard large-scale image dataset ImageNet \cite{deng2009imagenet}.
The recent work also implement the application of wavelet transform in deep network based image style transfer \cite{yoo2019photorealistic}.
Due to the absence of general wavelet transform layers,
the above methods are only evaluated using only one or two wavelets like Haar or dual-tree complex wavelet,
which is not extensively evaluated.

\section{Our method}
Our method is trying to design the general discrete wavelet transform (DWT/IDWT) layers
and apply them to improve the performance of CNNs for image classification.
We firstly present the basic wavelet theory.

\subsection{Wavelet}\label{subsec_wavelet}
\begin{table*}
	%\scriptsize
	\caption{Low-pass filters of the Daubechies wavelets.
    {\color{black}The high-pass filters of Daubechies wavelets could be deduced from the low-pass filters via Eq. (\ref{eq_high_pass_bank}).
    Daubechies wavelets are orthogonal.}}%, whose filters are applied to decompose and reconstruct image.}}
	\label{Tab_Daubechies_banks}
	\begin{center}
	\setlength{\tabcolsep}{1mm}{
	\begin{tabular}{c|cccccc}
		\hline
		          $p$           &     $1$      &       $2$       &         $3$         &         $4$         &         $5$         &         $6$         \\ \hline
		\multirow{12}{*}{$l_k$} &     $1$      &  $1+\sqrt{3}$   & $~~~0.332670552950$ & $~~~0.230377813309$ & $~~~0.160102397974$ & $~~~0.111540743350$ \\
		                        &     $1$      &  $3+\sqrt{3}$   & $~~~0.806891509311$ & $~~~0.714846570553$ & $~~~0.603829269797$ & $~~~0.494623890398$ \\
		                        &              &  $3-\sqrt{3}$   & $~~~0.459877502118$ & $~~~0.630880767930$ & $~~~0.724308528438$ & $~~~0.751133908021$ \\
		                        &              &  $1-\sqrt{3}$   &  $-0.135011020010$  &  $-0.027983769417$  & $~~~0.138428145901$ & $~~~0.315250351709$ \\
		                        &              &                 &  $-0.085441273882$  &  $-0.187034811719$  &  $-0.242294887066$  &  $-0.226264693965$  \\
		                        &              &                 & $~~~0.035226291886$ & $~~~0.030841381836$ &  $-0.032244869585$  &  $-0.129766867567$  \\
		                        &              &                 &                     & $~~~0.032883011667$ & $~~~0.077571493840$ & $~~~0.097501605587$ \\
		                        &              &                 &                     &  $-0.010597401785$  &  $-0.006241490213$  & $~~~0.027522865530$ \\
		                        &              &                 &                     &                     &  $-0.012580751999$  &  $-0.031582039317$  \\
		                        &              &                 &                     &                     & $~~~0.003335725285$ & $~~~0.000553842201$ \\
		                        &              &                 &                     &                     &                     & $~~~0.004777257511$ \\
		                        &              &                 &                     &                     &                     &  $-0.001077301085$  \\ \hline
		       $\text{factor}$         & $1/\sqrt{2}$ & $1/(4\sqrt{2})$ &         $1$         &         $1$         &         $1$         &         $1$         \\ \hline
	\end{tabular}}
	\end{center}
\end{table*}
\begin{table*}
%\scriptsize
\caption{Low-pass filters of the Cohen wavelets.
{\color{black}The high-pass filters of Cohen wavelets could be deduced via Eqs. (\ref{eq_high_pass_bank_bior_1})-(\ref{eq_high_pass_bank_bior_2}).
The filters and dual filters of biorthogonal Cohen wavelets are applied to decompose and reconstruct image, respectively.}}
\label{Tab_CDF_banks}
\begin{center}
	\setlength{\tabcolsep}{0.5mm}{
		\begin{tabular}{c|cc|cc|cc|cc}
			\hline
			$(p,\tilde{p})$     & \multicolumn{2}{|c|}{$(2,2)$}       & \multicolumn{2}{|c|}{$(3,3)$}       & \multicolumn{2}{|c|}{$(4,4)$}          & \multicolumn{2}{|c}{$(5,5)$}           \\ \hline
			filter          & $\textbf{\emph{l}}$ & $\tilde{\textbf{\emph{l}}}$ & $\textbf{\emph{l}}$ & $\tilde{\textbf{\emph{l}}}$ &  $\textbf{\emph{l}}$   & $\tilde{\textbf{\emph{l}}}$ &  $\textbf{\emph{l}}$   & $\tilde{\textbf{\emph{l}}}$ \\ \hline
			\multirow{12}{*}{$l_k$} &     $0$      &         $0$          &     $0$      &   $~~~0.06629126$    &       $0$       &         $0$          & $~~~0.01345671$ &         $0$          \\
			& $0.35355339$ &    $-0.17677670$     &     $0$      &    $-0.19887378$     &  $-0.06453888$  &   $~~~0.03782846$    &  $-0.00269497$  &         $0$          \\
			& $0.70710678$ &   $~~~0.35355339$    & $0.17677670$ &    $-0.15467961$     &  $-0.04068942$  &    $-0.02384947$     &  $-0.13670658$  &   $~~~0.03968709$    \\
			& $0.35355339$ &   $~~~1.06066017$    & $0.53033009$ &   $~~~0.99436891$    & $~~~0.41809227$ &    $-0.11062440$     &  $-0.09350470$  &   $~~~0.00794811$    \\
			&     $0$      &   $~~~0.35355339$    & $0.53033009$ &   $~~~0.99436891$    & $~~~0.78848562$ &   $~~~0.37740286$    & $~~~0.47680327$ &    $-0.05446379$     \\
			&     $0$      &    $-0.17677670$     & $0.17677670$ &    $-0.15467961$     & $~~~0.41809227$ &   $~~~0.85269868$    & $~~~0.89950611$ &   $~~~0.34560528$    \\
			&              &                      &     $0$      &    $-0.19887378$     &  $-0.04068942$  &   $~~~0.37740286$    & $~~~0.47680327$ &   $~~~0.73666018$    \\
			&              &                      &     $0$      &   $~~~0.06629126$    &  $-0.06453888$  &    $-0.11062440$     &  $-0.09350470$  &   $~~~0.34560528$    \\
			&              &                      &              &                      &       $0$       &    $-0.02384947$     &  $-0.13670658$  &    $-0.05446379$     \\
			&              &                      &              &                      &       $0$       &   $~~~0.03782846$    &  $-0.00269497$  &   $~~~0.00794811$    \\
			&              &                      &              &                      &                 &                      & $~~~0.01345671$ &   $~~~0.03968709$    \\
			&              &                      &              &                      &                 &                      &       $0$       &         $0$          \\ \hline
	\end{tabular}}
\end{center}
\end{table*}
Wavelet \cite{daubechies1992ten,mallat1989theory} is associated with scaling function $\phi(x)$ and wavelet function $\psi(x)$,
whose shifts and expansions compose stable basis for the signal space $L^2(x)$.
With the basis, a signal can be decomposed and reconstructed.
%Symmetric wavelet is with symmetric scaling function and symmetric/antisymmetric wavelet function,
%i.e., $\exists~a_0, a_1$ for $\forall~x$ satisfies $\phi(x) = \phi(x+a_0)$ and $\psi(x) = \pm\psi(x+a_1)$.
The scaling and wavelet functions of discrete wavelet are closely related with low-pass filter $\textbf{\emph{l}} = \{l_k\}_{k\in\mathbb{Z}}$
and high-pass filter $\textbf{\emph{h}} = \{h_k\}_{k\in\mathbb{Z}}$, respectively.
%$\phi(x), \psi(x)$ are symmetric/antisymmetric if and only if $\textbf{\emph{l}}, \textbf{\emph{h}}$ are symmetric/antisymmetric.
%It is not possible for any orthogonal wavelet with finite filters to be symmetric, except for the Haar.
In practice, these filters are applied for the data decomposition and reconstruction in DWT and IDWT.

\textbf{Orthogonal wavelets}\label{APP_ortho_wavelet}\quad
Daubechies wavelet is orthogonal, a set of orthogonal basis for $L^2(x)$ could be derived from its scaling and wavelet functions.
Daubechies wavelet has an approximation order parameter $p$,
and the length of its filter is $2p$.
Table \ref{Tab_Daubechies_banks} shows the low-pass filter $\textbf{\emph{l}} = \{l_k\}$ of the wavelets with order $p, 1\leq p\leq6$,
while the high-pass filter $\textbf{\emph{h}} = \{h_k\}$ can be deduced from
\begin{equation}\label{eq_high_pass_bank}
h_k = (-1)^k l_{\mathcal{N}-k},
\end{equation}
where $\mathcal{N}$ is an odd number.
Daubechies(1) is Haar wavelet.

\textbf{Biorthogonal wavelets}\label{APP_bior_wavelet}\quad
Cohen wavelets are symmetric biorthogonal wavelets,
and each of them is associated with scaling function $\phi$, wavelet function $\psi$, and their dual functions $\tilde{\phi}, \tilde{\psi}$.
Correspondingly, it has four filters $\textbf{\emph{l}}$, $\textbf{\emph{h}}$, $\tilde{\textbf{\emph{l}}}$, and $\tilde{\textbf{\emph{h}}}$.
While a signal is decomposed using filters $\textbf{\emph{l}}$ and $\textbf{\emph{h}}$ with DWT,
it can be reconstructed using the dual filters $\tilde{\textbf{\emph{l}}}$ and $\tilde{\textbf{\emph{h}}}$ with IDWT.
Cohen wavelet is with two order parameters $p$ and $\tilde{p}$.
Table \ref{Tab_CDF_banks} shows the low-pass filters with orders $2\leq p = \tilde{p}\leq5$.
Their high-pass filters can be deduced from
\begin{eqnarray}
\label{eq_high_pass_bank_bior_1}
&h_k = (-1)^k \tilde{l}_{\mathcal{N}-k},&\\
\label{eq_high_pass_bank_bior_2}
&\tilde{h}_k = (-1)^k l_{\mathcal{N}-k},&
\end{eqnarray}
where $\mathcal{N}$ is an odd number.
Cohen$(1,1)$ is Haar wavelet.

Wavelet theory is valid for finite or infinite filters,
but the infinite case is rarely covered in practical interest.
%Discrete Wavelet Transform (DWT) decomposes an image into its low-frequency and high-frequency components.
%In theory, Inverse DWT (IDWT) could precisely reconstructs the image using the DWT output.
%In practice, in our DWT and IDWT layers, we truncate \textbf{L}, \textbf{H} to fit the image with finite size.
%As a result, IDWT with asymmetric wavelet can not fully restore the image in the region near the image boundary,
%and the region width increases as the wavelet filter length increases.
%With symmetric wavelet, however, one can fully restore the image based on symmetric extension of the DWT output.
%In other words, the DWT output of symmetric wavelets save the fully information of the input image,
%but that of asymmetric wavelets do not.
%We believe that is the reason why asymmetric Daubechies wavelets perform worse
%than symmetric Cohen and Haar wavelets in image classification of WaveCNets.
%Our DWT and IDWT layers are applicable to any orthogonal or biorthogonal wavelets.
Besides the othogonal and biorthogonal wavelets,
various wavelets and beyond-wavelets, including
multi-wavelets \cite{Keinert2004Wavelets}, dual-tree complex wavelet \cite{selesnick2005the}, ridgelet \cite{jianwei2006combined},
curvelet \cite{Cand2006Fast}, bandelet \cite{pennec2000image}, and contourlet \cite{do2005the}, etc.,
have been designed and studied.
Wavelets have been widely used in signal processing, numerical analysis,
pattern recognition, computer vision, quantum mechanics, etc.
%The DWT/IDWT layer could be applicable to these mathematical tools with slight modification.

\subsection{Wavelet transform layers}\label{subsec_wavelet_transform_layers}
The key issues for the general wavelet transform (DWT/IDWT) layers are the data forward and backward propagations.
Although the following analysis is for orthogonal wavelets and 1D/2D data,
it can be generalized to other discrete wavelets and 3D data with only slight changes.

\subsubsection{1D DWT/IDWT}
DWT decomposes a given 1D data $\textbf{\emph{x}} = \{x_j\}_{j\in\mathbb{Z}}$
into its low-frequency component $\textbf{\emph{x}}_{\textrm{low}} = \{x_k^{(\textrm{low})}\}_{k\in\mathbb{Z}}$
and high-frequency component $\textbf{\emph{x}}_{\textrm{high}} = \{x_k^{(\textrm{high})}\}_{k\in\mathbb{Z}}$,
where
\begin{equation}\label{eq_DWT}
\left\{
\begin{array}{l}
x_k^{(\textrm{low})} = \sum_j l_{j-2k} x_j, \\
x_k^{(\textrm{high})} = \sum_j h_{j-2k} x_j,
\end{array}
\right.
\end{equation}
and $\textbf{\emph{l}} = \{l_k\}_{k\in\mathbb{Z}}, \textbf{\emph{h}} = \{h_k\}_{k\in\mathbb{Z}}$ are the low-pass
and high-pass filters of an orthogonal wavelet.
According to Eq. (\ref{eq_DWT}), DWT consists of filtering and down-sampling.
IDWT reconstructs $\textbf{\emph{x}}$ using $\textbf{\emph{x}}_{\textrm{low}}, \textbf{\emph{x}}_{\textrm{high}}$, where
\begin{equation}\label{eq_IDWT}
x_j = \sum_k\left(l_{j-2k} x_k^{(\textrm{low})} + h_{j-2k} x_k^{(\textrm{high})}\right).
\end{equation}

In expressions with matrices and vectors, Eq. (\ref{eq_DWT}) and Eq. (\ref{eq_IDWT}) can be rewritten as
\begin{align}
\label{eq_DWT_1D_L}
&\textbf{\emph{x}}_{\textrm{low}} = \mathcal {L}\,\textbf{\emph{x}},\\
\label{eq_DWT_1D_H}
&\textbf{\emph{x}}_{\textrm{high}} = \mathcal {H}\,\textbf{\emph{x}},\\
\label{eq_IDWT_1D}
&\textbf{\emph{x}} = \mathcal {L}^T \textbf{\emph{x}}_{\textrm{low}} + \mathcal {H}^T \textbf{\emph{x}}_{\textrm{high}},
\end{align}
where
\begin{align}
%\nonumber to remove numbering (before each equation)
\mathcal {L} &=
\left(
\begin{array}{ccccccc}
	\cdots & \cdots & \cdots &        &        &        &                \\
	\cdots & l_{-1} &  l_0   &  l_1   & \cdots &        &                \\
	       &        & \cdots & l_{-1} &  l_0   &  l_1   & \cdots         \\
	       &        &        &        &        & \cdots & \cdots
\end{array}
\right),\\
%\end{align}
%\begin{align}
\mathcal {H} &=
\left(
\begin{array}{ccccccc}
	\cdots & \cdots & \cdots &        &        &        &          \\
	\cdots & h_{-1} &  h_0   &  h_1   & \cdots &        &          \\
	       &        & \cdots & h_{-1} &  h_0   &  h_1   & \cdots   \\
	       &        &        &        &        & \cdots & \cdots
\end{array}
\right).
\end{align}

Eqs. (\ref{eq_DWT_1D_L}) -- (\ref{eq_IDWT_1D}) present the forward propagations for 1D DWT and IDWT.
The backward propagation of 1D DWT is closely associated with the gradients
$\frac{\partial \textbf{\emph{x}}_{\textrm{low}}}{\partial\textbf{\emph{x}}}$ and $\frac{\partial \textbf{\emph{x}}_{\textrm{high}}}{\partial\textbf{\emph{x}}}$,
which can be derived from Eqs. (\ref{eq_DWT_1D_L}),
\begin{equation}\label{eq_DWT_bp}
\frac{\partial \textbf{\emph{x}}_{\textrm{low}}}{\partial\textbf{\emph{x}}} = \mathcal {L}^T,~~~~
\frac{\partial\textbf{\emph{x}}_{\textrm{high}}}{\partial\textbf{\emph{x}}} = \mathcal {H}^T.
\end{equation}
Similarly, for the back propagation of IDWT, the key issues are the gradients
$\frac{\partial\textbf{\emph{x}}}{\partial\textbf{\emph{x}}_{\textrm{low}}}, \frac{\partial\textbf{\emph{x}}}{\partial\textbf{\emph{x}}_{\textrm{high}}}$,
which can be derived from Eq. (\ref{eq_IDWT_1D}),
\begin{equation}\label{eq_IDWT_bp}
\frac{\partial\textbf{\emph{x}}}{\partial\textbf{\emph{x}}_{\textrm{low}}} = \mathcal {L},~~~~
\frac{\partial\textbf{\emph{x}}}{\partial\textbf{\emph{x}}_{\textrm{high}}} = \mathcal {H}.
\end{equation}

\subsubsection{2D DWT/IDWT}\label{subsubsec_2D_DWT_IDWT}
Given a 2D data $\textbf{\emph{X}}$, the DWT usually do 1D DWT
on its every row and column, i.e.,
\begin{align}
\label{eq_DWT_2D_ll}
\textbf{\emph{X}}_{ll} &= \mathcal {L} \textbf{\emph{X}} \mathcal {L}^T, \\
\label{eq_DWT_2D_lh}
\textbf{\emph{X}}_{lh} &= \mathcal {H} \textbf{\emph{X}} \mathcal {L}^T, \\
\label{eq_DWT_2D_hl}
\textbf{\emph{X}}_{hl} &= \mathcal {L} \textbf{\emph{X}} \mathcal {H}^T,\\
\label{eq_DWT_2D_hh}
\textbf{\emph{X}}_{hh} &= \mathcal {H} \textbf{\emph{X}} \mathcal {H}^T,
\end{align}
and the corresponding 2D IDWT is implemented with
\begin{equation}\label{eq_DWT_2D_M}
\textbf{\emph{X}} = \mathcal{L}^T \textbf{\emph{X}}_{ll} \mathcal{L}
+ \mathcal{H}^T \textbf{\emph{X}}_{lh} \mathcal{L}
+ \mathcal{L}^T \textbf{\emph{X}}_{hl} \mathcal{H}
+ \mathcal{H}^T \textbf{\emph{X}}_{hh} \mathcal{H}.
\end{equation}

In the output of 2D DWT, $\textbf{\emph{X}}_{ll}$ is the low-frequency component of input $\textbf{\emph{X}}$,
which represents the main information, including the basic object structures;
$\textbf{\emph{X}}_{lh}, \textbf{\emph{X}}_{hl}, \textbf{\emph{X}}_{hh}$ are three high-frequency components,
which save the horizontal, vertical, and diagonal details of $\textbf{\emph{X}}$, respectively.

Suppose 2D DWT/IDWT are applied in deep network,
then the backward propagation of 2D DWT can be implemented by the gradients,
$\frac{\partial \textbf{\emph{X}}_{ll}}{\partial\textbf{\emph{X}}},
\frac{\partial \textbf{\emph{X}}_{lh}}{\partial\textbf{\emph{X}}},
\frac{\partial \textbf{\emph{X}}_{hl}}{\partial\textbf{\emph{X}}},
\frac{\partial \textbf{\emph{X}}_{hh}}{\partial\textbf{\emph{X}}},$
\begin{align}
\label{eq_ll_lh_dwt}
\frac{\partial \textbf{\emph{X}}_{ll}}{\partial\textbf{\emph{X}}}(G) &= \mathcal {L}^T G\,\mathcal {L},\\
\frac{\partial \textbf{\emph{X}}_{lh}}{\partial\textbf{\emph{X}}}(G) &= \mathcal {H}^T G\,\mathcal {L},\\
\label{eq_hl_hh_dwt}
\frac{\partial \textbf{\emph{X}}_{hl}}{\partial\textbf{\emph{X}}}(G) &= \mathcal {L}^T G\,\mathcal {H},\\
\frac{\partial \textbf{\emph{X}}_{hh}}{\partial\textbf{\emph{X}}}(G) &= \mathcal {H}^T G\,\mathcal {H},
%\vspace{-15pt}
\end{align}
where $G$ is the backward propagation output from the layer following the 2D DWT.

Similarly, the backward propagation of 2D IDWT is implemented by gradients,
$\frac{\partial \textbf{\emph{X}}}{\partial\textbf{\emph{X}}_{ll}},
\frac{\partial \textbf{\emph{X}}}{\partial\textbf{\emph{X}}_{lh}},
\frac{\partial \textbf{\emph{X}}}{\partial\textbf{\emph{X}}_{hl}},
\frac{\partial \textbf{\emph{X}}}{\partial\textbf{\emph{X}}_{hh}},$
\begin{align}
\label{eq_ll_lh_idwt}
&\frac{\partial \textbf{\emph{X}}}{\partial\textbf{\emph{X}}_{ll}}(G) = \mathcal {L}\,G\,\mathcal {L}^T,\\
&\frac{\partial \textbf{\emph{X}}}{\partial\textbf{\emph{X}}_{lh}}(G) = \mathcal {H}\,G\,\mathcal {L}^T,\\
\label{eq_hl_hh_idwt}
&\frac{\partial \textbf{\emph{X}}}{\partial\textbf{\emph{X}}_{hl}}(G) = \mathcal {L}\,G\,\mathcal {H}^T,\\
&\frac{\partial \textbf{\emph{X}}}{\partial\textbf{\emph{X}}_{hh}}(G) = \mathcal {H}\,G\,\mathcal {H}^T,\vspace{-8pt}
\end{align}
where $G$ is the backward propagation output from the layer following the 2D IDWT.

The forward and backward propagations of 3D DWT and IDWT are slightly more complicated, but similar to that of 1D/2D DWT and IDWT.
In practice, we choose the wavelets associated with finite filters, as that shown in Table \ref{Tab_Daubechies_banks} and Table \ref{Tab_CDF_banks}.
%Haar wavelet with $\textbf{\emph{l}} = \frac{1}{\sqrt{2}}\{1,1\}$ and $\textbf{\emph{h}} = \frac{1}{\sqrt{2}}\{1,-1\}$.
For finite data $\textbf{\emph{x}} \in \mathbb{R}^N$ and  $\textbf{\emph{X}} \in \mathbb{R}^{M\times N}$,
the $\mathcal {L}, \mathcal {H}$ are truncated to be the size of $\lfloor\frac{N}{2}\rfloor\times N$ or $\lfloor\frac{M}{2}\rfloor\times M$.

We rewrite 1D/2D/3D DWT and IDWT as network layers in PyTorch \cite{paszke2017automatic},
which are applicable to various discrete orthogonal and biorthogonal wavelets, like Haar, Daubechies, and Cohen.
In the layers, we do DWT and IDWT channel by channel for multi-channel data.

\subsection{WaveCNets}
\begin{figure}[bpt]
	\centering
	\subfigure[The general denoising approach using wavelet.]{
		\label{fig_denoise_a}
		\includegraphics*[scale=0.7, viewport=265 308 585 400]{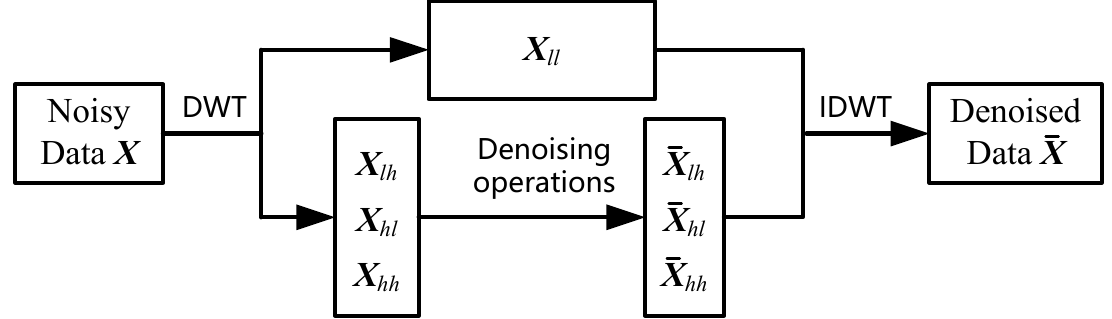}
	}\\
	\subfigure[The simplest wavelet based ``denoising'' method, $\text{DWT}_{ll}$.]{
		\label{fig_denoise_b}
		\includegraphics*[scale=0.7, viewport=265 228 585 258]{figures/Visio-down_sampling_of_dwt.pdf}
	}
	\caption{The general denoising approach based on wavelet transforms and the one used in WaveCNet.
            {\color{black}While the general denoising approach keeps the data size, the simplest ``denoising'' method, $\text{DWT}_{ll}$, halves the data size.}}
	\label{fig_dual_structures}
\end{figure}

The noise are mostly represented by the high-frequency components in a noisy 2D data $\textbf{\emph{X}}$.
Therefore, as Fig. \ref{fig_denoise_a} shows, the general wavelet based denoising \cite{donoho1995noising,donoho1994ideal}
consists of three steps:
(1) decompose the noisy data $\textbf{\emph{X}}$ using DWT into low-frequency component $\textbf{\emph{X}}_{ll}$
and high-frequency components $\textbf{\emph{X}}_{lh}, \textbf{\emph{X}}_{hl}, \textbf{\emph{X}}_{hh}$,
(2) filter the high-frequency components,
(3) reconstruct the data with the processed components using IDWT.
In this paper, we choose the simplest wavelet based ``denoising'', i.e., dropping the high-frequency components,
as Fig. \ref{fig_denoise_b} shows.
$\text{DWT}_{ll}$ denotes the transform mapping the data to its low-frequency component.

\begin{figure}[t]
	\centering
	\includegraphics*[scale=0.85, viewport=25 698 295 785]{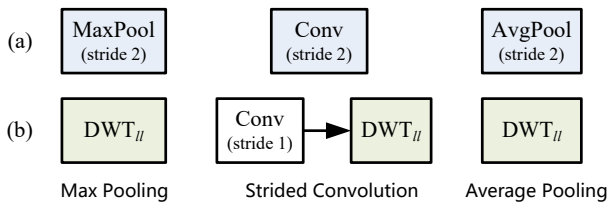}
	\caption{(a) Baseline, the commonly used down-sampling operations in deep networks.
	(b) Wavelet integrated down-sampling in WaveCNets.
    {\color{black}While the commonly used down-sampling operations may suffer from aliasing effects,
    our wavelet integrated down-sampling do low-pass filter before down-sampling,
    which helps suppressing the aliasing effects.}
	}\label{fig_down_sampling_of_dwt}
\end{figure}

Based on the commonly used CNNs, including VGG16bn, ResNets, and DenseNet121,
we design WaveCNets (WVGG16bn, WResNets, WDenseNet121) by replacing their down-sampling operation with $\text{DWT}_{ll}$.
As Fig. \ref{fig_down_sampling_of_dwt} shows, WaveCNets replace the max-pooling and average-pooling with $\text{DWT}_{ll}$,
and upgrade strided-convolution using convolution with stride of $1$ followed by $\text{DWT}_{ll}$,
i.e.,
\begin{align}
\label{eq_maxpool_up}
\text{MaxPool}_{s = 2} \rightarrow &\ \text{DWT}_{ll},\\
\label{eq_conv_up}
\text{Conv}_{s = 2} \rightarrow &\ \text{DWT}_{ll} \circ \text{Conv}_{s=1},\\
\label{eq_avgpool_up}
\text{AvgPool}_{s = 2} \rightarrow &\ \text{DWT}_{ll},
\end{align}
where ``$\text{MaxPool}_s$'', ``$\text{Conv}_s$'', and ``$\text{AvgPool}_s$'' denote the max-pooling,
strided-convolution, and average-pooling with stride $s$, respectively.

In the down-sampling of WaveCNets, while $\text{DWT}_{ll}$ halves the size of the feature maps,
it denoises them by removing their high-frequency components.
The output of $\text{DWT}_{ll}$, i.e., the low-frequency component, saves the main information of the feature map
to extract the identifiable features.
In other words, $\text{DWT}_{ll}$ could help WaveCNets to suppress the aliasing effect,
i.e., maintain the basic object structures in the the CNN feature maps and resist the noise propagation.
Therefore, it is expected that wavelet will %accelerate the CNN training and
lead to higher accuracy and better noise-robustness for CNN based image classification.

\subsection{Computation for DWT/IDWT}\label{subsec_computation_analysis}
Compared with the original CNNs, WaveCNets do not employ new learnable parameters,
while the wavelet transform introduces additional computation.

We here analyze the multiply-add operations increased by wavelet transform in WaveCNets.
Given a 2D tensor $\textbf{\emph{X}}$ with size of $M\times N$ and $C$ channels,
the amount of multiply-adds used in 2D DWT is
\begin{align}
\label{eq_no_2D_DWT}
4C\left(M^2N+\dfrac{MN^2}{2} - \dfrac{3MN}{4}\right),
\end{align}
and the amount of multiply-adds used in 2D IDWT is
\begin{align}
\label{eq_no_2D_IDWT}
4C\left(MN^2+\dfrac{M^2N}{2} - \dfrac{3MN}{4}\right) + 3,
\end{align}
according to Eqs. (\ref{eq_DWT_1D_L}) -- (\ref{eq_DWT_2D_hh}).

Table \ref{Tab_ratio_wavelet_convolution} presents
the ratios of wavelet related multiply-adds over the total operations for WaveCNets,
when the input size is $3\times224\times224$.
We only count the amount of multiply-adds in $\text{DWT}_{ll}$ for WaveCNets.
\begin{table}
%\scriptsize
%\small
\caption{Multiply-add operation numbers in WaveCNets.
{\color{black}DWT$_{ll}$ are complicated compared with the common down-sampling in the original CNNs,
leading to more multiply-add operations.}}
\label{Tab_ratio_wavelet_convolution}
\begin{center}
\begin{threeparttable}
\setlength{\tabcolsep}{1.25mm}{
\begin{tabular}{r||c|cccc|c}\hline
\multirow{3}{*}{}	&\multicolumn{6}{c}{WaveCNet}\\\cline{2-7}
							&VGG & \multicolumn{4}{c|}{ResNet}&DenseNet \\\cline{2-7}
							&16bn&18&34&50&101&121\\\hline
baseline\tnote{*} ($\times10^9$)&    $15.51$&     $1.82$& $3.67$ &$4.12$&$7.85$ & $2.88$\\\hline
non-wavelet ($\times10^9$)&    $15.51$&     $1.82$& $3.67$ &$4.12$&$7.85$ & $2.88$\\
wavelet ($\times10^9$)&    $1.43$&\multicolumn{4}{c|}{$0.22$}&$0.18$\\\cdashline{1-7}[2pt/2pt]
ratio ($\%$)   &    $8.44$&     $10.85$&    $5.69$& $5.11$& $2.75$&     $5.82$\\\hline
\end{tabular}}
\begin{tablenotes}
	\item[*] {\color{black}corresponding to the multiply-add operation numbers of original CNNs (i.e., VGG16bn, ResNets, DenseNet121),
    which are almost the same with the non-wavelet operation numbers in the WaveCNets.}
\end{tablenotes}
\end{threeparttable}
\end{center}
\end{table}

\section{Experiments}
\begin{table*}[t]
	%\scriptsize
	%\small
	\caption{Top-1 accuracy of WaveCNets on ImageNet validation set.
    {\color{black}The symmetric Haar and Cohen wavelets improve the accuracy of CNNs (VGG16bn, ResNets, and DenseNet121),
    while the asymmetric Daubechies wavelets (``db4'', ``db5'', and ``db6'') with high approximation orders may decrease the CNN performance.}}
	\label{Tab_WaveCNet_accuracy}
	\begin{center}
	\begin{threeparttable}
		\setlength{\tabcolsep}{2.25mm}{
			\begin{tabular}{cc||c|cccc|c}\hline
		\multicolumn{2}{c||}{Wavelet}&{WVGG16bn} &WResNet18&WResNet34&WResNet50&WResNet101 &{WDenseNet121} \\\hline\hline
\multicolumn{2}{c||}{none (baseline)\tnote{*}}
							&73.37		     &69.76		&73.30		&76.15		&77.37		&74.65		\\\hline
\multicolumn{2}{c||}{Haar}
	 						&74.10 ($+$0.73) &71.47 ($+$1.71)	&74.35 ($+$1.05)	&\textbf{76.89 ($+$0.74)}&78.23 ($+$0.86)	&75.27 ($+$0.62)	\\\hline
{\multirow{4}{*}{Cohen}}
&\multicolumn{1}{|c||}{ch2.2}&74.31 ($+$0.94)&\textbf{71.62 ($+$1.86)}	&74.33 ($+$1.03)	&76.41 ($+$0.26)&78.34 ($+$0.97)	&75.36 ($+$0.71)	\\
&\multicolumn{1}{|c||}{ch3.3}&\textbf{74.40 ($+$1.03)}&71.55 ($+$1.79)	&74.51 ($+$1.21)	&76.71 ($+$0.56)&\textbf{78.51 ($+$1.14)}	&\textbf{75.44 ($+$0.79)}\\
&\multicolumn{1}{|c||}{ch4.4}&74.02 ($+$0.65)&71.52 ($+$1.76)	&\textbf{74.61 ($+$1.31)}	&76.56 ($+$0.41)&78.47 ($+$1.10)	&75.29 ($+$0.64)	\\
&\multicolumn{1}{|c||}{ch5.5}&73.67 ($+$0.30)&71.26 ($+$1.50)	&74.34 ($+$1.04)	&76.51 ($+$0.36)	&78.39 ($+$1.02)	&75.01 ($+$0.36)	\\\hline
{\multirow{5}{*}{Daubechies}}&
\multicolumn{1}{|c||}{db2} 	 &74.08 ($+$0.71)&71.48 ($+$1.72)	&74.30 ($+$1.00)	&76.27 ($+$0.12)	&78.29 ($+$0.92)	&75.08 ($+$0.43)	\\
&\multicolumn{1}{|c||}{db3}	 &73.64 ($+$0.27)&71.08 ($+$1.32)	&74.11 ($+$0.81)	&76.38 ($+$0.23)	&77.80 ($+$0.43)	&74.66 ($+$0.01)		\\
&\multicolumn{1}{|c||}{db4}	 &72.71	($-$0.66)&70.35 ($+$0.59)	&73.53 ($+$0.23)	&75.65 ($-$0.50)	&76.85 ($-$0.52)	&73.68 ($-$0.97)		\\
&\multicolumn{1}{|c||}{db5}	 &71.15	($-$2.22)&69.54 ($-$0.22)	&73.41 ($+$0.11)	&74.90 ($-$1.25)	&76.19 ($-$1.18)	&72.35 ($-$2.30)		\\
&\multicolumn{1}{|c||}{db6}	 &69.06	($-$4.31)&68.74 ($-$1.02)	&72.68 ($-$0.62)	&73.95 ($-$2.20)	&75.36 ($-$2.01)	&70.99 ($-$3.66)		\\\hline
			\end{tabular}}
		\begin{tablenotes}
			%\tiny
			\item[*] corresponding to the results of original CNNs, i.e., VGG16bn, ResNets, DenseNet121.
		\end{tablenotes}
	\end{threeparttable}
	\end{center}
\vspace{-0pt}
\end{table*}

In our experiments, we mainly evaluate classification accuracies and noise-robustness of WaveCNets
using ImageNet \cite{deng2009imagenet}.
We also test the improvement of WaveCNets on the classification of adversarial samples and detection of COCO.
%At last, we explore the potential of wavelet integrated deep networks for image segmentation.

\subsection{ImageNet classification}
ImageNet contains 1.2M training and 50K validation images from 1000 categories.
WaveCNets are trained on the training set from scratch when various wavelets are used.
We train them 90 epochs using stochastic gradient descent (SGD) with batch size of $256$.
The initial learning rate is $0.05$ for WVGG16bn and $0.1$ for WResNets and WDenseNet121,
which is decayed by multiplying $0.1$ every $30$ epochs.

Table \ref{Tab_WaveCNet_accuracy} presents the top-1 accuracy of WaveCNets on ImageNet validation set,
where ``haar'', ``db$p$'', and ``ch$p.\tilde{p}$'' denote the Haar wavelet, Daubechies wavelet with approximation order $p$,
and Cohen wavelet with orders $(p,\tilde{p})$.
%The length of the wavelet filters increase as the orders increase.
While Haar and Cohen wavelets are symmetric, Daubechies are not.
%The supplementary materials present the filters of these wavelets.

In Table \ref{Tab_WaveCNet_accuracy}, the classification results of the original CNNs are taken as the baseline,
which are sourced from the official PyTorch\footnote{https://pytorch.org/docs/stable/torchvision/models.html\#classification}.
The accuracy difference of WaveCNets compared with the baseline are parenthesized in the table.
The symmetric Haar and Cohen wavelets improve the classification accuracy of all CNNs,
while the best wavelet varies with CNN.
Take ResNet18 for example, the symmetric wavelets improve the accuracy by about $1.50\%$,
and the best performance ($71.62\%$) of WResNet18 is achieved with wavelet ``ch2.2''.
However, as the order increases, the asymmetric Daubechies wavelet decreases the classification accuracy of CNNs.
Daubechies wavelets with lower orders (``db2'' and ``db3'') could improve the CNN accuracy,
while that with higher orders (``db5'' and ``db6'') may reduce the accuracy.
For example, the top-1 accuracy of WResNet18 decreases from $71.48\%$ to $68.74\%$, as the order increases from $2$ to $6$.
In general, CNNs transformed by the symmetric wavelets perform better than that by asymmetric wavelets.
%Therefore, we do not train WVGG16bn, WResNet101, and WDenseNet121 using Daubechies wavelets with order higher than $2$.

\begin{figure}[bpt]
	\centering
	\includegraphics*[scale=0.625, viewport=21 2 417 305]{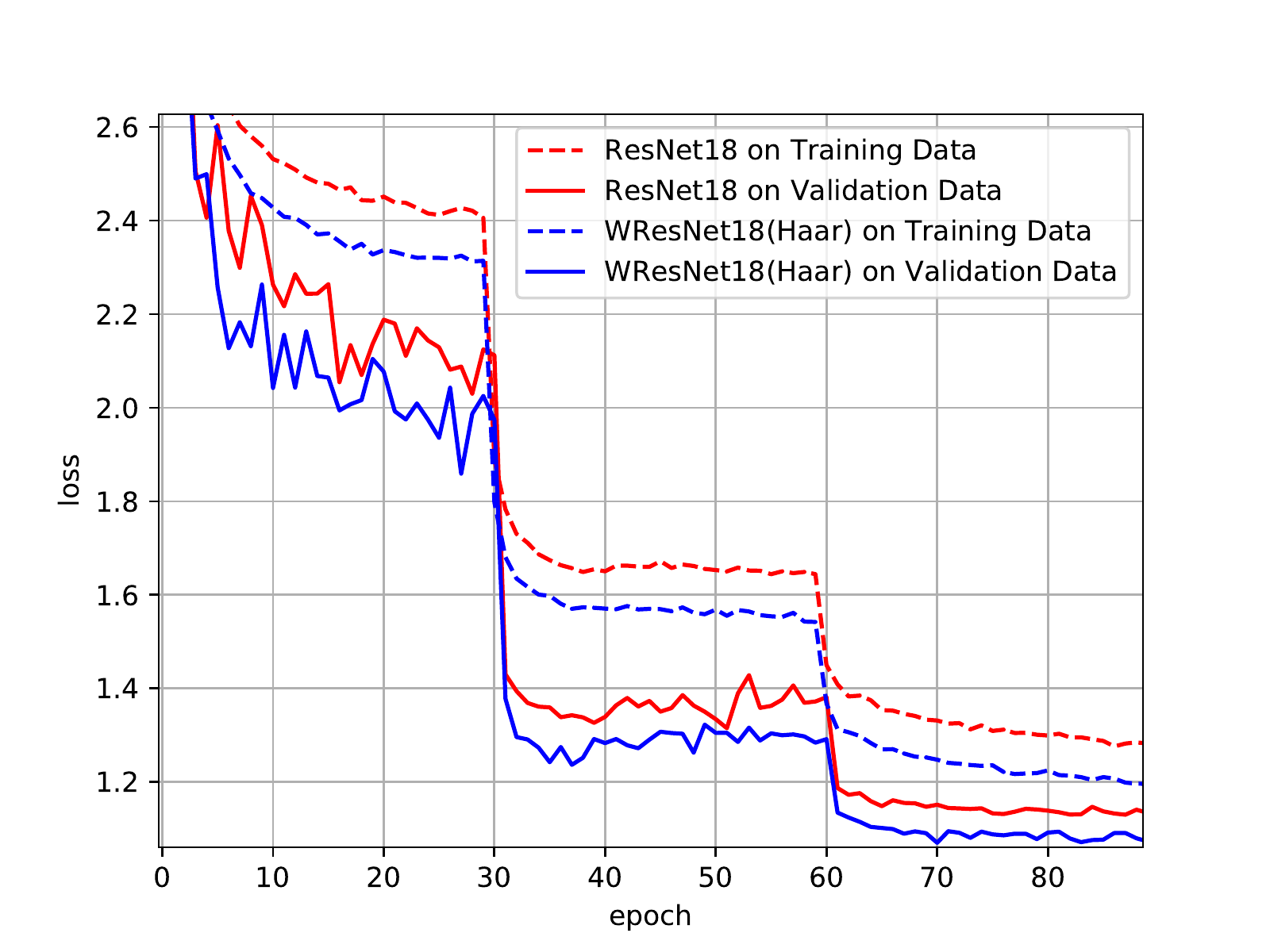}
	\caption{The loss of ResNet18 and WResNet18(Haar).
    {\color{black}In their first 10 epochs, the loss of WResNet18(Haar) decreases faster than that of ResNet18,
    and the tread keeps nearly the same in the following epochs,
    which suggests that wavelet could accelerates the CNN training in the initial training stage.}}
	\label{fig_loss_resnet18}
\end{figure}
\begin{figure*}[bpt]
	\centering\small
	%\subfigure[\emph{stupa}]
	%{\includegraphics*[scale=0.215, viewport=200 102 968 606]{figures/feature_map_clean_new/WVGG16bn_n04346328_00011836.pdf}\label{fig_feature_maps_a}}\hspace{5pt}
	\subfigure[\emph{suit}]
	{\includegraphics*[scale=0.215, viewport=200 102 968 606]{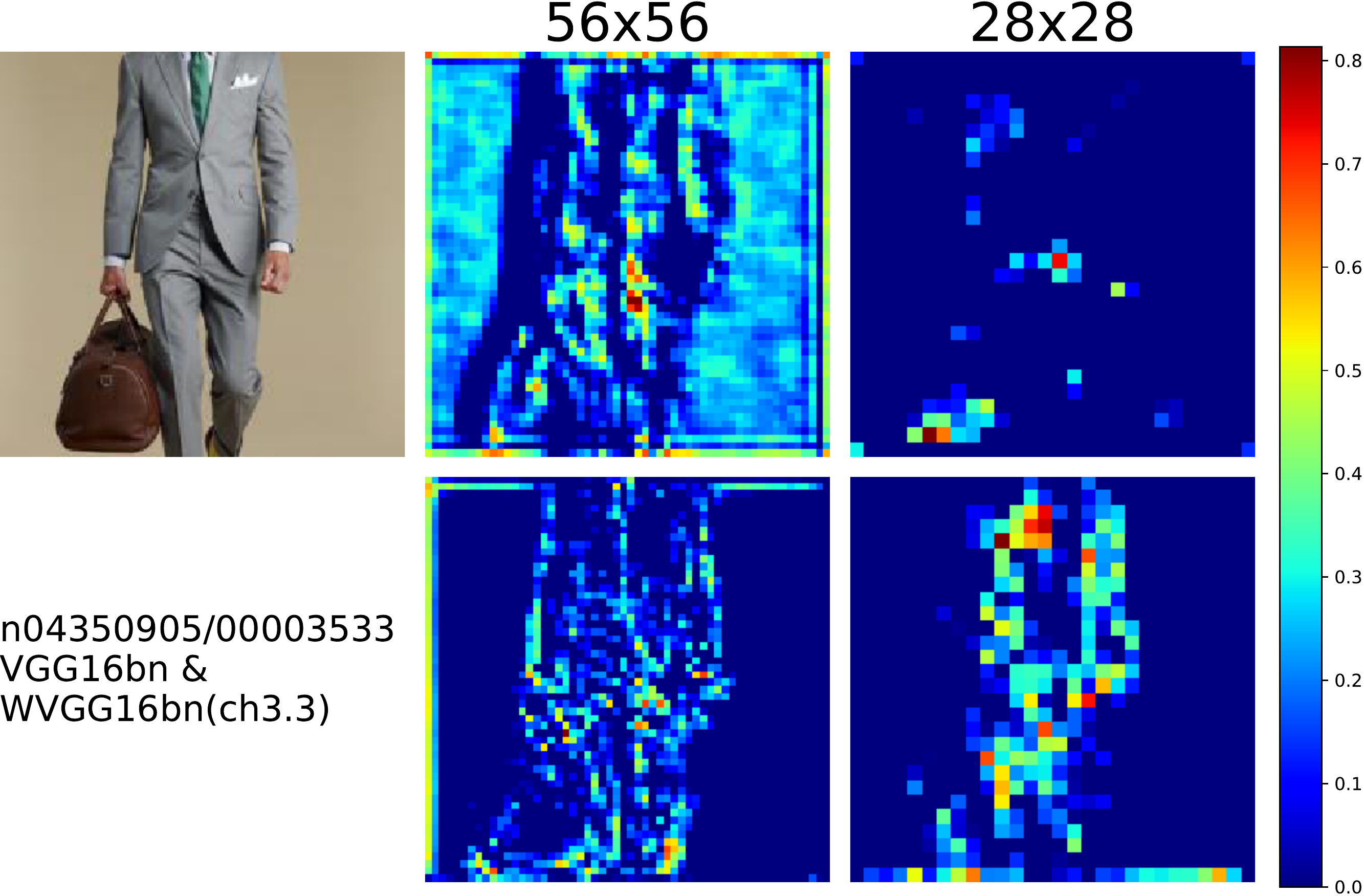}\label{fig_feature_maps_b}}\hspace{5pt}
	%\subfigure[\emph{espresso}]
	%{\includegraphics*[scale=0.155, viewport=200 102 968 606]{figures/feature_map_clean_new/WVGG16bn_n07920052_00028898.pdf}\label{fig_feature_maps_c}}\hspace{5pt}
	\subfigure[\emph{espresso}]
	{\includegraphics*[scale=0.215, viewport=200 102 968 606]{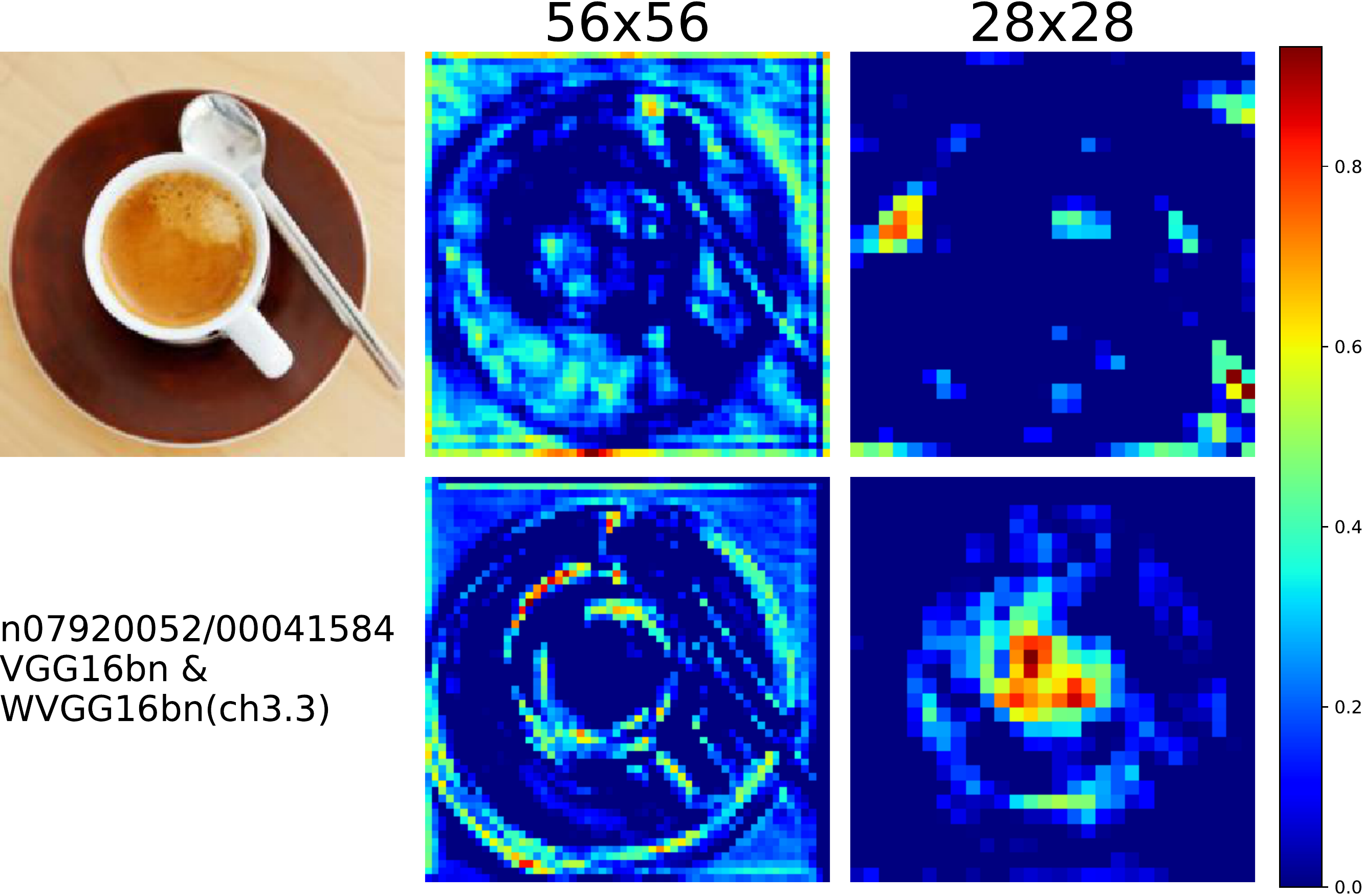}\label{fig_feature_maps_d}}\hspace{5pt}
	%\subfigure[\emph{English springer}]
	%{\includegraphics*[scale=0.215, viewport=200 102 968 606]{figures/feature_map_clean_new/ResNet18_n02102040_00010994.pdf}\label{fig_feature_maps_e}}\hspace{5pt}
	\subfigure[\emph{lycaenid}]
	{\includegraphics*[scale=0.215, viewport=200 102 968 606]{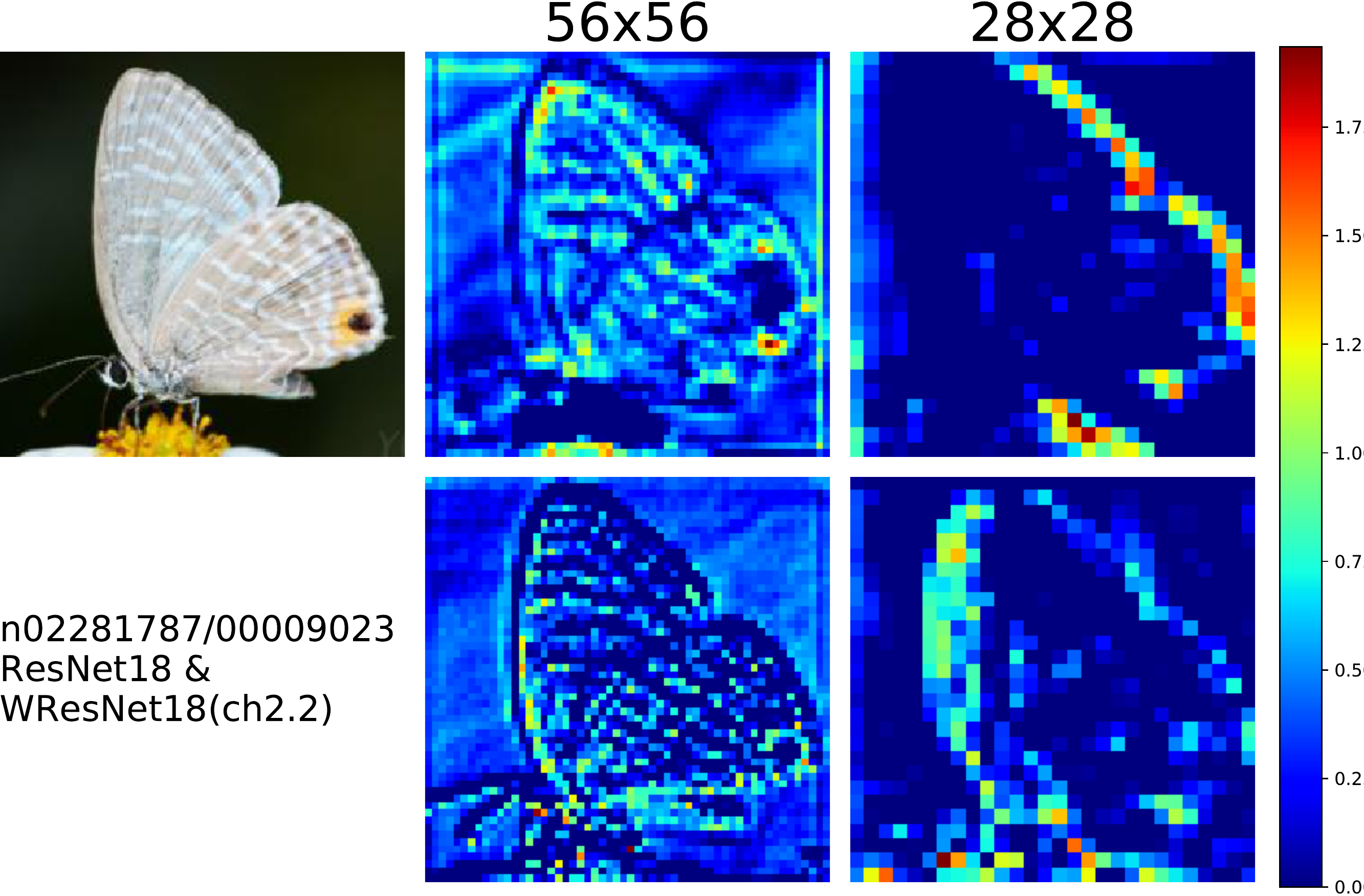}\label{fig_feature_maps_f}}\hspace{0pt}\\
	\subfigure[\emph{palace}]
	{\includegraphics*[scale=0.215, viewport=200 102 968 606]{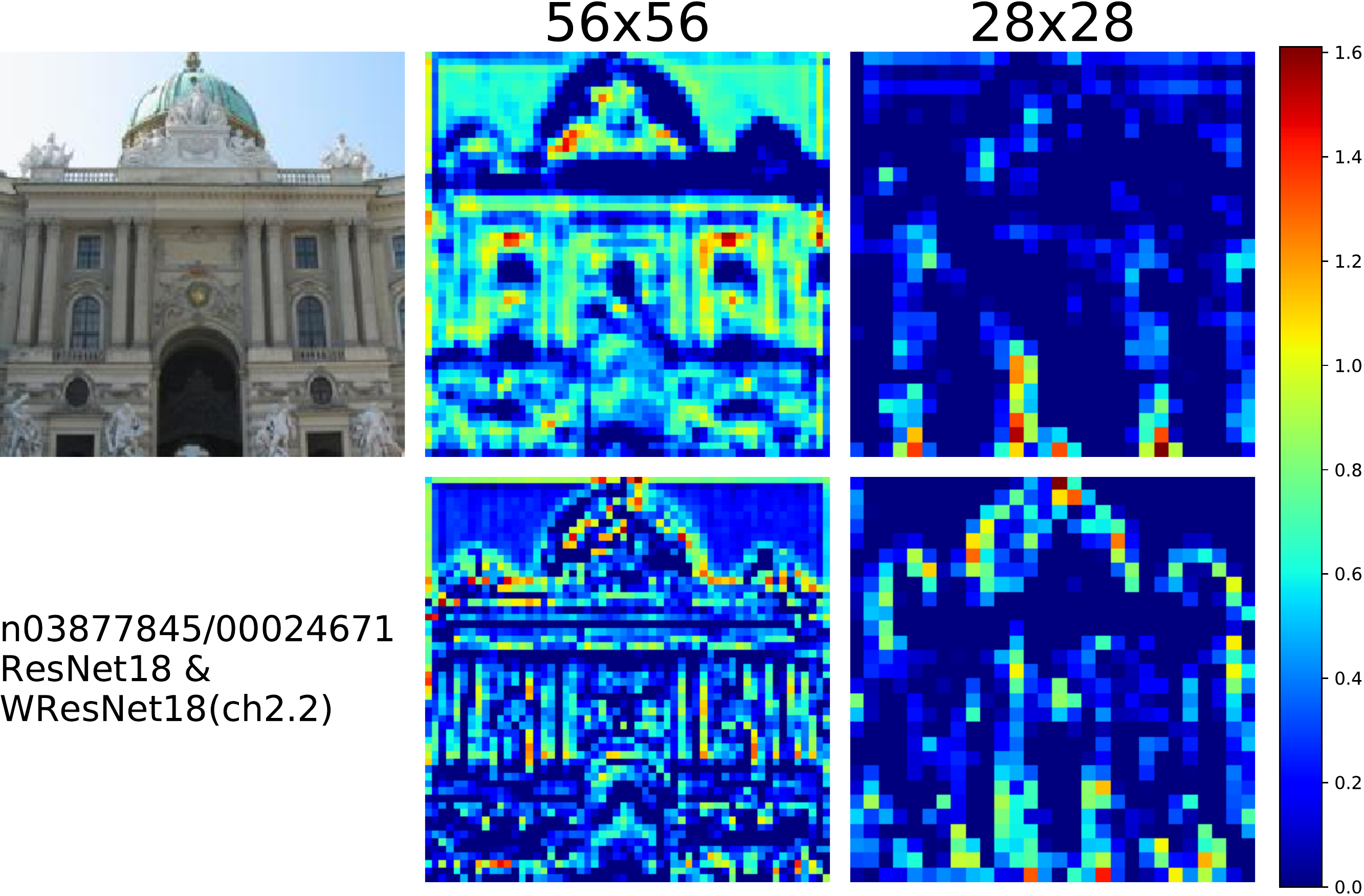}\label{fig_feature_maps_g}}\hspace{5pt}
	%\subfigure[\emph{vase}]
	%{\includegraphics*[scale=0.155, viewport=200 102 968 606]{figures/feature_map_clean_new/ResNet18_n04522168_00031441.pdf}\label{fig_feature_maps_h}}\hspace{0pt}\\
	%\subfigure[\emph{jay}]
	%{\includegraphics*[scale=0.155, viewport=200 102 968 606]{figures/feature_map_clean_new/ResNet34_n01580077_00026465.pdf}\label{fig_feature_maps_i}}\hspace{5pt}
	\subfigure[\emph{jay}]
	{\includegraphics*[scale=0.215, viewport=200 102 968 606]{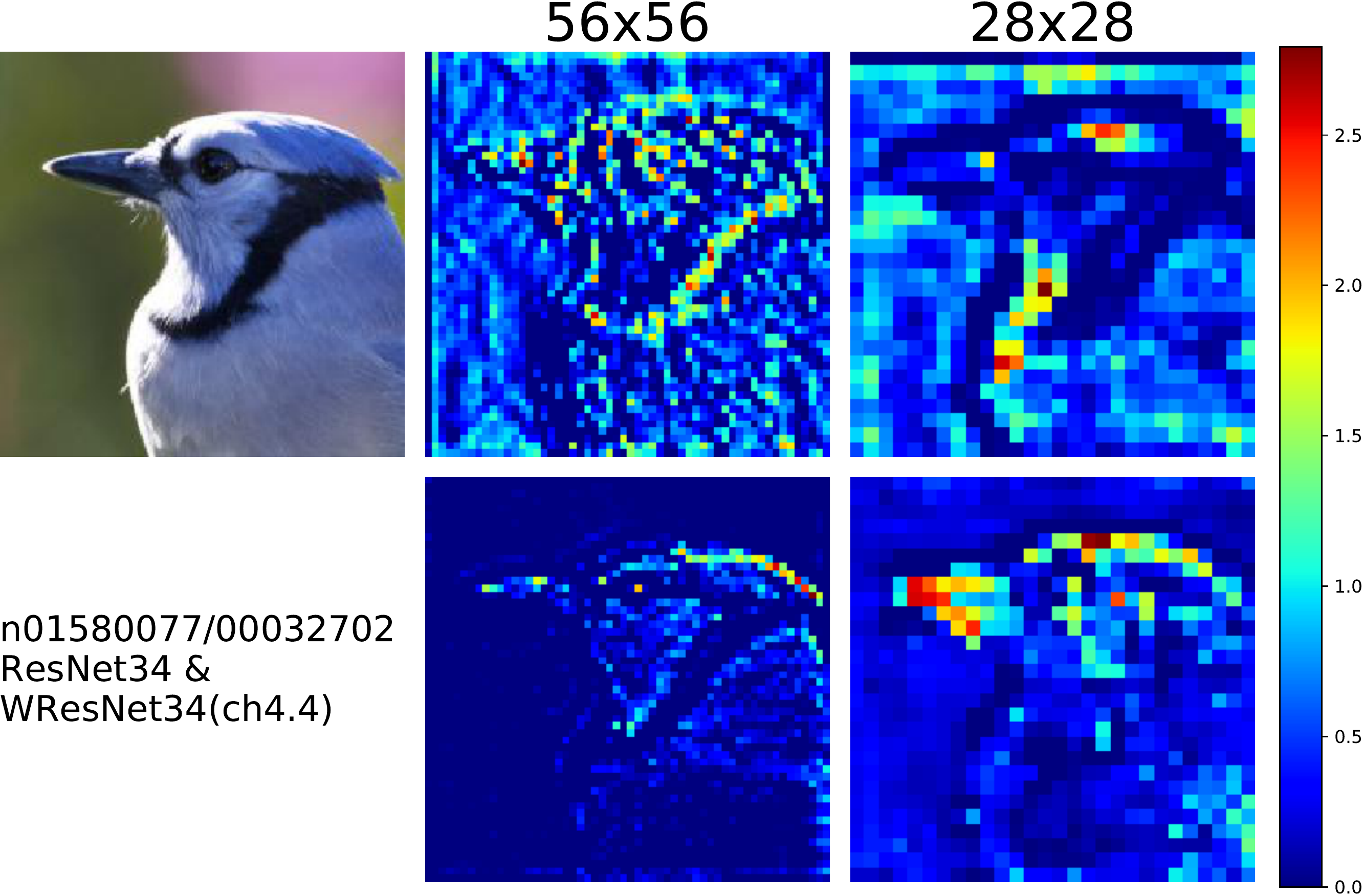}\label{fig_feature_maps_j}}\hspace{5pt}
	%\subfigure[\emph{minivan}]
	%{\includegraphics*[scale=0.155, viewport=200 102 968 606]{figures/feature_map_clean_new/ResNet34_n03770679_00007374.pdf}\label{fig_feature_maps_k}}\hspace{5pt}
	\subfigure[\emph{minivan}]
	{\includegraphics*[scale=0.215, viewport=200 102 968 606]{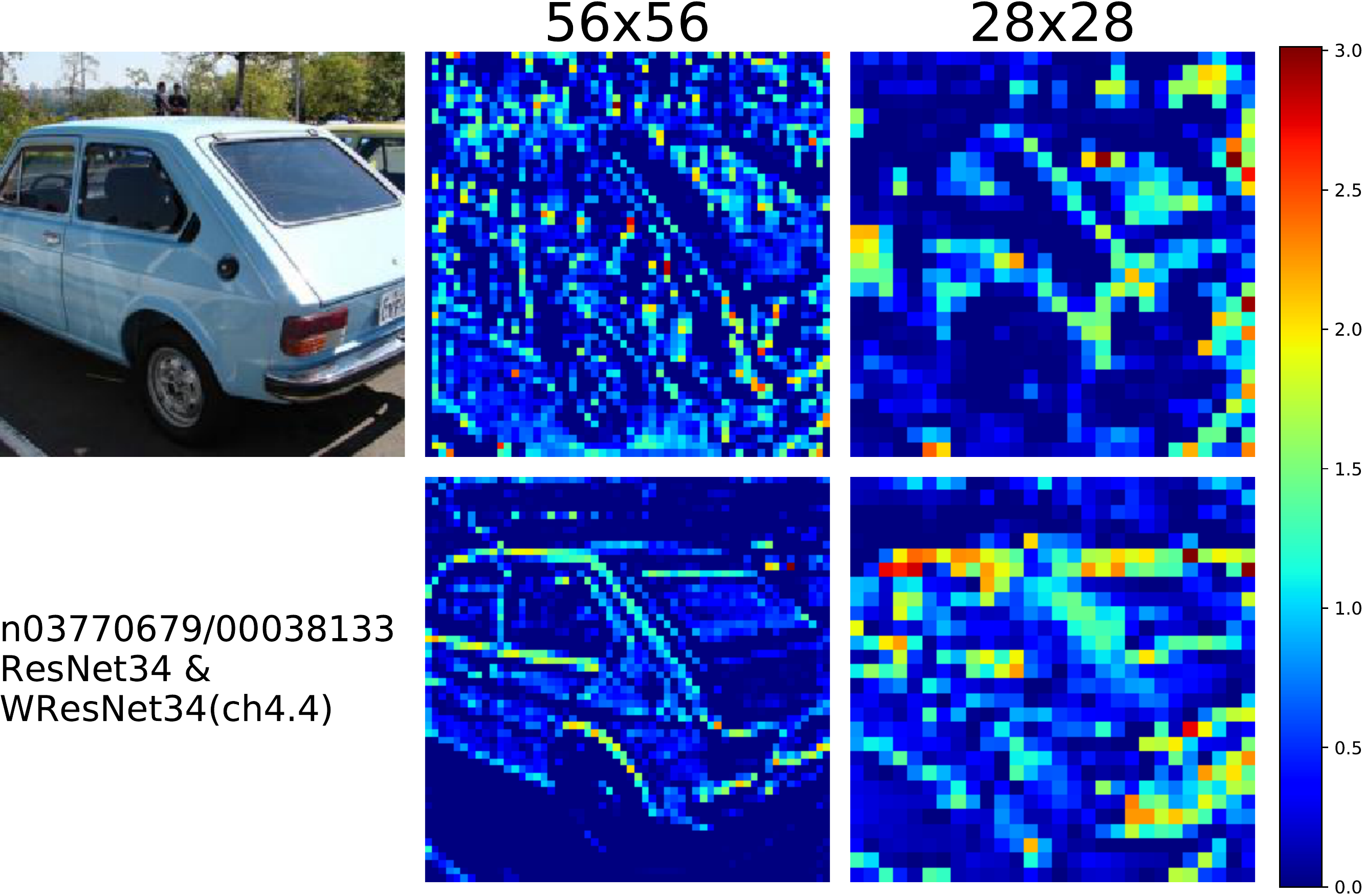}\label{fig_feature_maps_l}}\hspace{5pt}\\
	%\subfigure[\emph{quilt}]
	%{\includegraphics*[scale=0.215, viewport=200 102 968 606]{figures/feature_map_clean_new/ResNet34_n04033995_00009944.pdf}\label{fig_feature_maps_m}}\hspace{0pt}\\
	%\subfigure[\emph{teapot}]
	%{\includegraphics*[scale=0.215, viewport=200 102 968 606]{figures/feature_map_clean_new/ResNet34_n04398044_00019622.pdf}\label{fig_feature_maps_n}}\hspace{5pt}
	\subfigure[\emph{wall clock}]
	{\includegraphics*[scale=0.215, viewport=200 102 968 606]{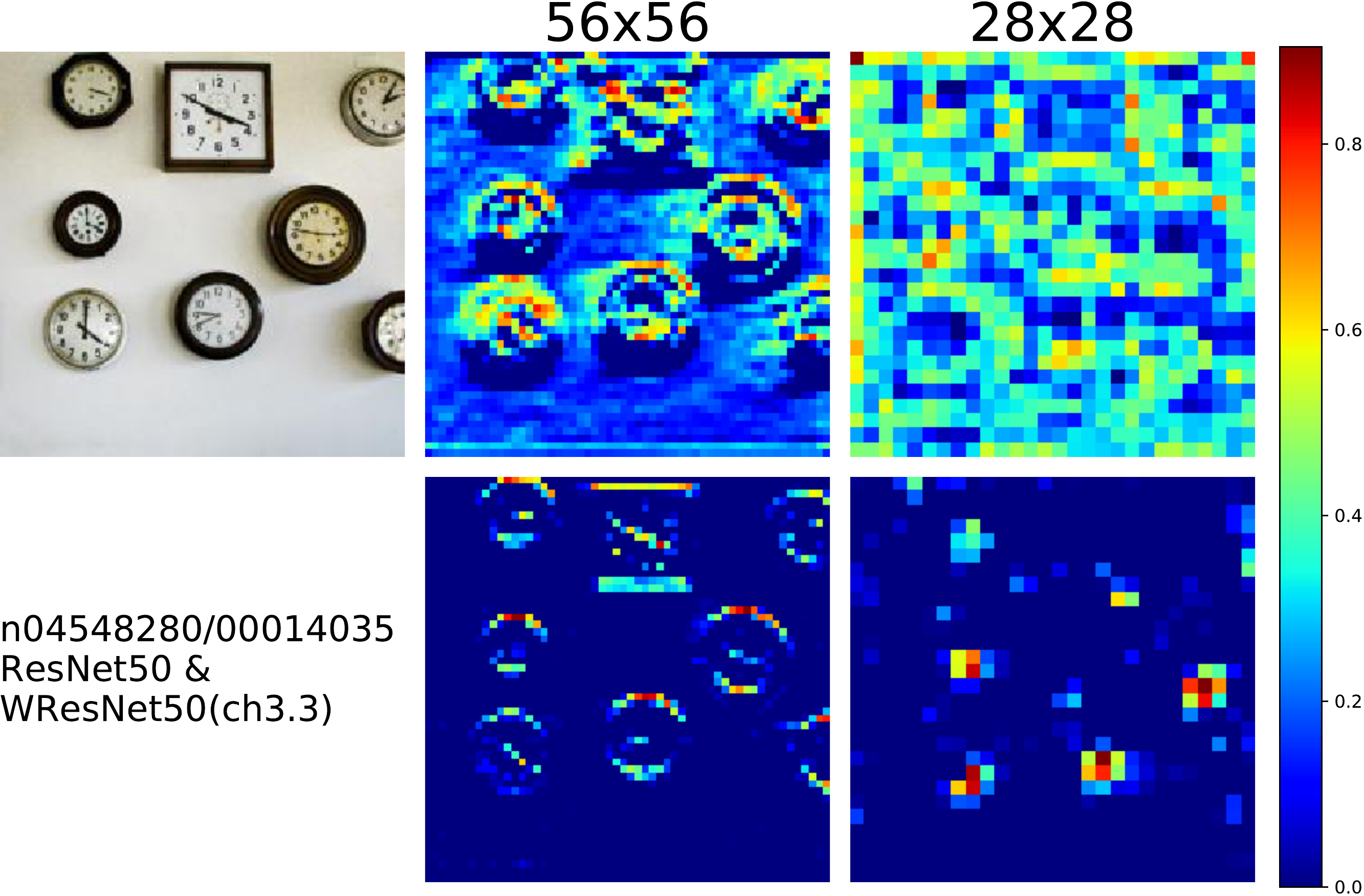}\label{fig_feature_maps_o}}\hspace{5pt}
	%\subfigure[\emph{wall clock}]
	%{\includegraphics*[scale=0.215, viewport=200 102 968 606]{figures/feature_map_clean_new/ResNet50_n04548280_00018936.pdf}\label{fig_feature_maps_p}}\hspace{0pt}\\
	%\subfigure[\emph{harp}]
	%{\includegraphics*[scale=0.215, viewport=200 102 968 606]{figures/feature_map_clean_new/ResNet101_n03495258_00039792.pdf}\label{fig_feature_maps_q}}\hspace{5pt}
	\subfigure[\emph{schooner}]
	{\includegraphics*[scale=0.215, viewport=200 102 968 606]{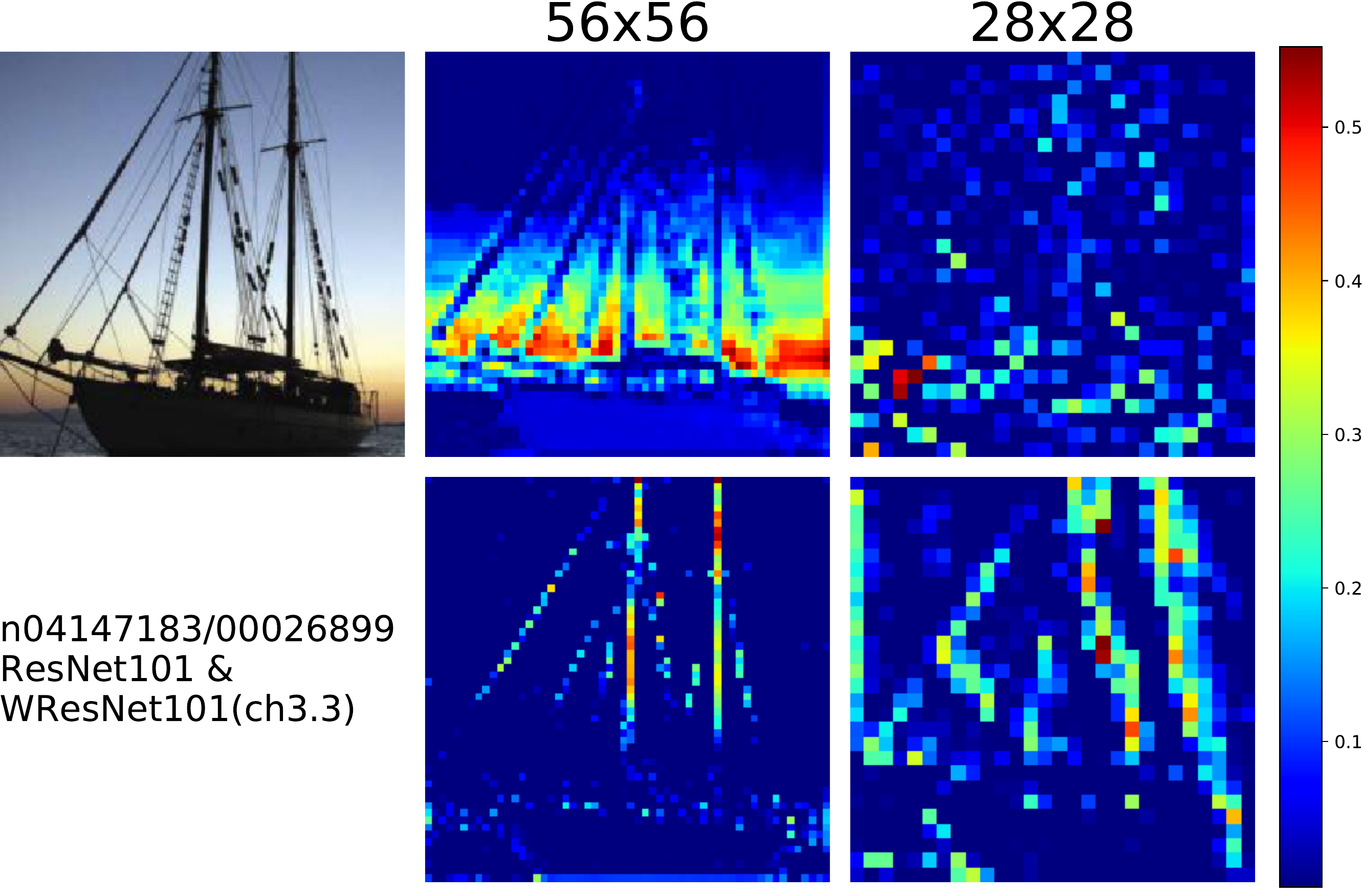}\label{fig_feature_maps_r}}\hspace{5pt}
	\subfigure[\emph{stupa}]
	{\includegraphics*[scale=0.215, viewport=200 102 968 606]{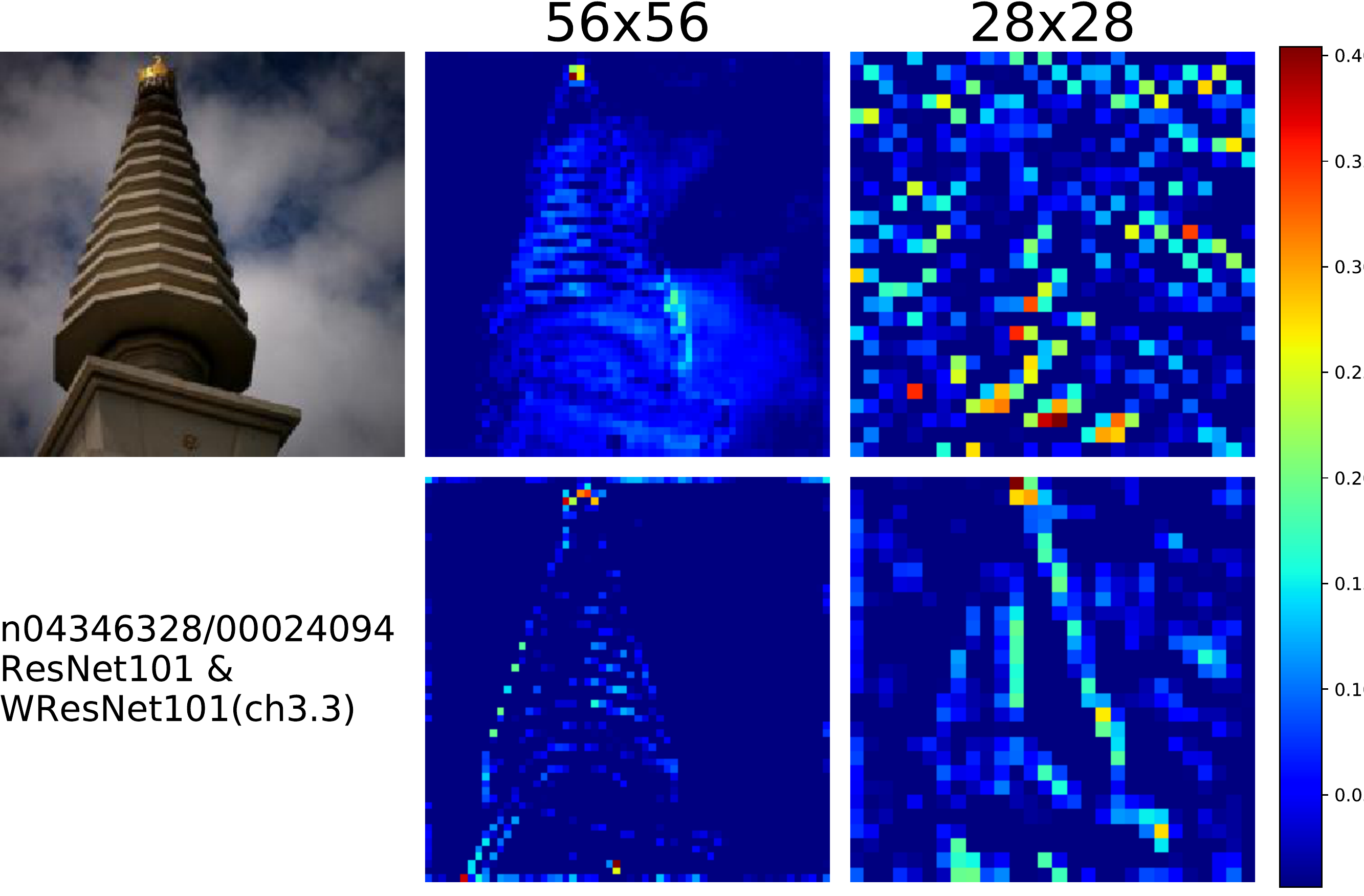}\label{fig_feature_maps_s}}\hspace{0pt}
	%\subfigure[\emph{espresso}]
	%{\includegraphics*[scale=0.155, viewport=200 102 968 606]{figures/feature_map_clean_new/ResNet101_n07920052_00041584.pdf}\label{fig_feature_maps_t}}\hspace{0pt}\\
	\caption{The feature maps of CNNs (top) and WaveCNets (bottom).
    {\color{black}In each subfigure, the first row shows the input image and its two feature maps output from the original CNNs,
    while the second row shows the related information (the image, CNN, and WaveCNet names) and the feature maps output from the WaveCNets.
    Compared with CNNs, the feature maps of WaveCNets are cleaner and the object structures are more complete.
    %The feature maps of WaveCNets are cleaner than that of CNNs, and the object structures in the former are more complete than that in the latter.
    }}
	\label{fig_feature_maps}
\end{figure*}
We retrain the original ResNet18 using the standard ImageNet classification training repository in PyTorch.
Fig. \ref{fig_loss_resnet18} compares the losses of ResNet18 and WResNet18(Haar) during the training procedure,
which adopts red dashed and blue dashed lines to denote the train losses of ResNet18 and WResNet18(Haar), respectively.
{\color{black}In the initial training stage (the first 10 epochs), the loss of WResNet18(Haar) decreases faster than that of ResNet18,
and the trends become similar in the following epochs, which suggests that wavelet accelerates the ResNet18 training in the initial training stage.}
Finally, the training loss of WResNet18(Haar) is about $0.08$ lower than that of ResNet18,
although they employ the same amount of learnable parameters.
On the validation set, WResNet18 loss (blue solid line) is also always lower than ResNet18 loss (red solid line),
which leads to the increase of classification accuracy by $1.71\%$.

Fig. \ref{fig_feature_maps} presents nine example feature maps produced by well trained CNNs and WaveCNets for the ImageNet validation images.
%More examples are shown in supplementary materials.
In each subfigure, the first row shows the input image with size of $224\times224$
and its two feature maps produced by original CNN,
while the second row shows the related information (image, CNN, and WaveCNet names) and the feature maps produced by the WaveCNet.
The two feature maps are captured from the 16th output channel of the final layer
in the network blocks with tensor size of $56\times56$ (middle) and $28\times28$ (right), respectively.
We enlarge and color the feature maps for better illustration.

The examples in Fig. \ref{fig_feature_maps} cover the various deep network architectures
(VGG16bn, ResNet18, ResNet34, ResNet50, and ResNet101) and various objects
(\emph{suit}, \emph{espresso}, \emph{lycaenid}, \emph{palace}, \emph{jay}, \emph{minican}, \emph{wall clock}, \emph{stupa}, etc.).
While the original CNNs adopt various down-sampling operations,
including max-pooling, average-pooling, and strided-convolution, WaveCNets replace them using DWT$_{ll}$.
From Fig. \ref{fig_feature_maps}, one can find
that the backgrounds of the feature maps produced by WaveCNets are cleaner than that produced by CNNs,
and the object structures in the former are more complete than that in the latter.
For example, in the first row of Fig. \ref{fig_feature_maps_o},
the \emph{wall clock} boundary in the ResNet50 feature map with size of $56\times56$ are fuzzy,
and the basic structures of \emph{wall clocks} have been totally hidden by strong noise in the feature map with size of $28\times28$.
In the second row,
the backgrounds of feature maps produced by WResNet50(ch3.3) are very clean,
and it is easy to figure out the \emph{wall clock} structures in the feature map with size of $56\times56$
and the \emph{wall clock} areas in the feature map with size of $28\times28$.
The above observations illustrate that the common down-sampling operations could result in aliasing effect, i.e., accumulating noise and breaking the basic object structures,
while DWT in WaveCNets relieves the drawbacks.
We believe that this is the reason why WaveCNets achieve increased accuracy.

In \cite{zhang2019making}, the author is surprised at the increased classification accuracy of CNNs
after low-pass filtering is integrated into the down-sampling.
In \cite{geirhos2018imagenet},
the authors show that ``ImageNet-trained CNNs are strongly biased towards recognising textures rather than shapes''.
Our experimental results suggest that this may be sourced from the commonly used down-sampling operations,
which tend to break the object structures and accumulate noise in the feature maps.

\subsection{Noise-robustness}
In \cite{hendrycks2019benchmarking},
the authors corrupt the ImageNet validation set using 15 visual corruptions with five severity levels,
to create ImageNet-C and test the robustness of ImageNet-trained classifiers to the input corruptions.
The 15 corruptions are sourced from four categories,
i.e., noise (Gaussian noise, shot noise, impulse noise),
blur (defocus blur, frosted glass blur, motion blur, zoom blur),
weather (snow, frost, fog, brightness), and digital (contrast, elastic, pixelate, JPEG-compression).
$E_{s,c}^f$ denotes the top-1 error of a trained classifier $f$ on corruption type $c$ at severity level $s$.
The authors present the Corruption Error $\text{CE}_c^f$, computed with
\begin{align}
\label{eq_CE}
\text{CE}_c^f &= \sum_{s=1}^5 E_{s,c}^f\left/\sum_{s=1}^5 E_{s,c}^{\text{AlexNet}}\right.,
\end{align}
to evaluate the performance of a trained classifier $f$.
In Eq. (\ref{eq_CE}), the authors normalize the error using the top-1 error of AlexNet \cite{krizhevsky2012imagenet}
to adjust the difference of various corruptions.

In this section, we use the noise part (750K images, 50K $\times$ 3 $\times$ 5) of ImageNet-C
and
\begin{align}
\label{eq_mCE_noise}
\text{mCE}_{\text{noise}}^f &= \dfrac{1}{3}\left(\text{CE}_{\text{Gaussian}}^f
+ \text{CE}_{\text{shot}}^f + \text{CE}_{\text{impulse}}^f\right)
\end{align}
to evaluate the noise-robustness of WaveCNet $f$.

\begin{figure}[bpt]
	\centering
	\includegraphics*[scale=0.55, viewport=31 18 471 447]{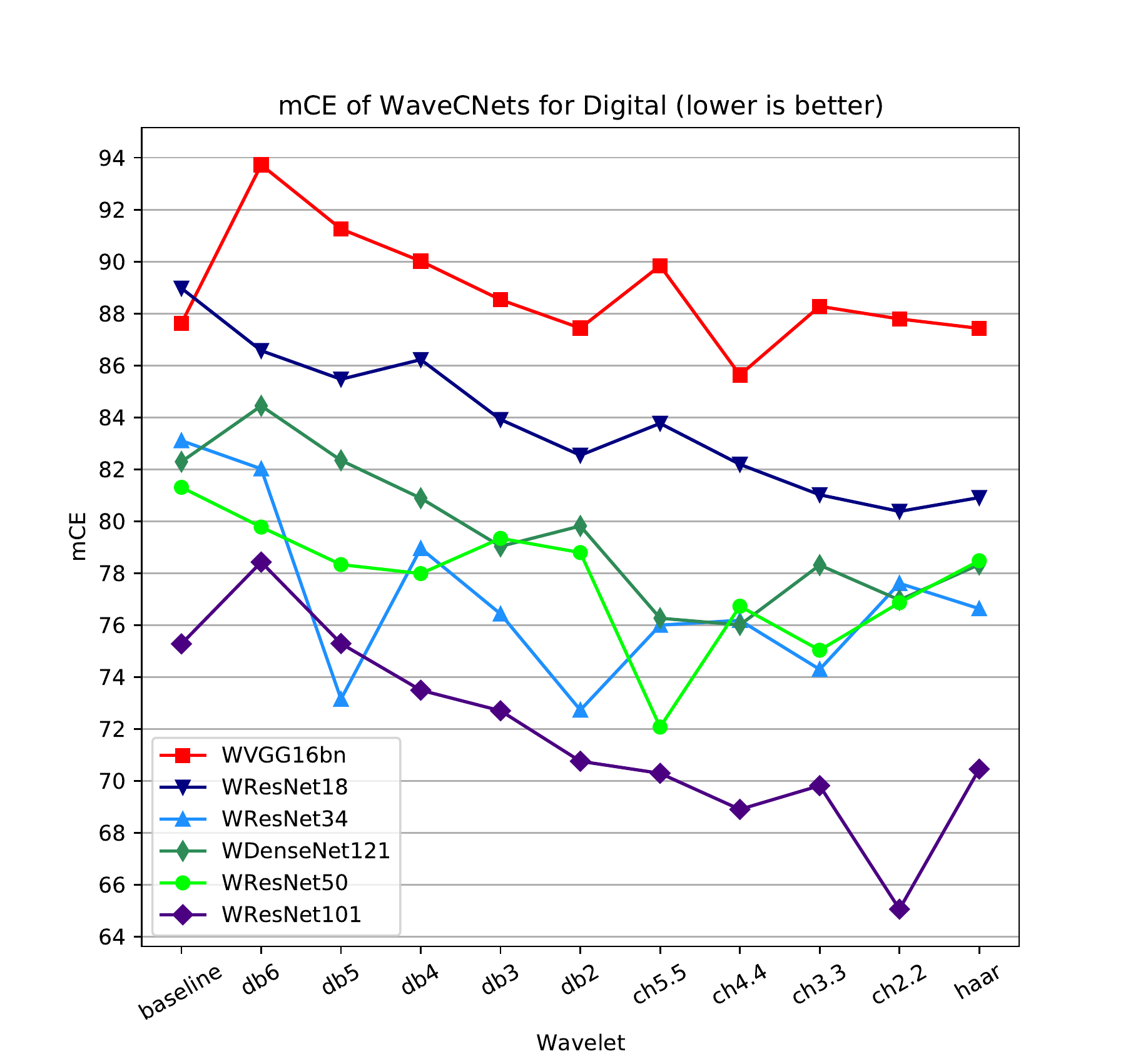}
	\caption{The noise mCE of WaveCNets.
    {\color{black}In the figure, the ``baseline'' corresponds to the noise mCEs of original CNNs,
    while ``db$p$'', ``ch$p.\tilde{p}$'' and ``haar'' correspond to the mCEs of WaveCNets with different wavelets.
    Except VGG16bn, our method obviously increase the noise-robustness of the CNNs for image classification.}}
	\label{fig_mCE_noise}
\end{figure}
\begin{figure*}[bpt]
	\centering
	\subfigure[\emph{vase}]
	{\includegraphics*[scale=0.5, viewport=182 68 657 640]{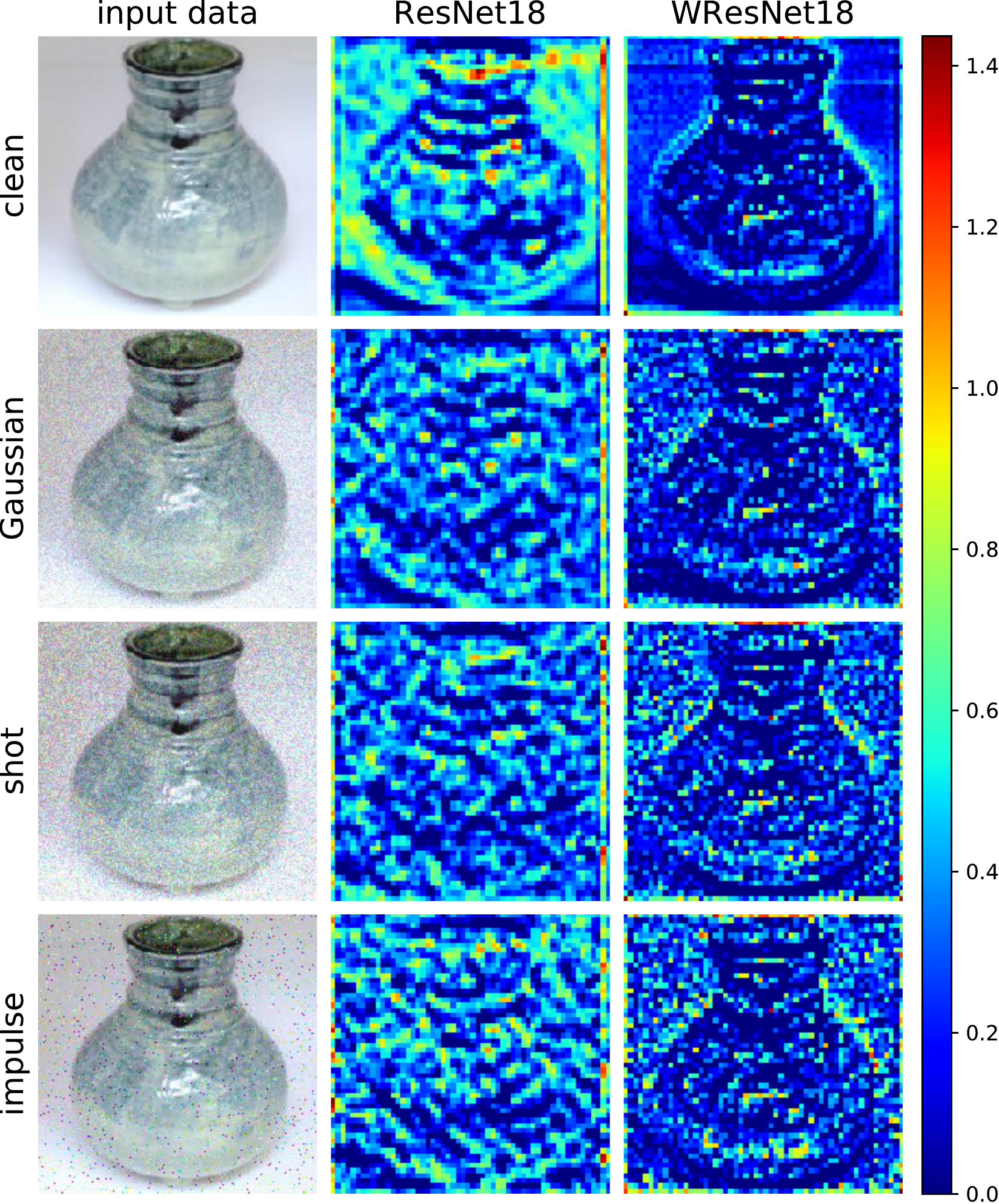}
	\label{fig_feature_map_noise_a}}\hspace{20pt}
	\subfigure[\emph{moped}]
	{\includegraphics*[scale=0.5, viewport=182 68 657 640]{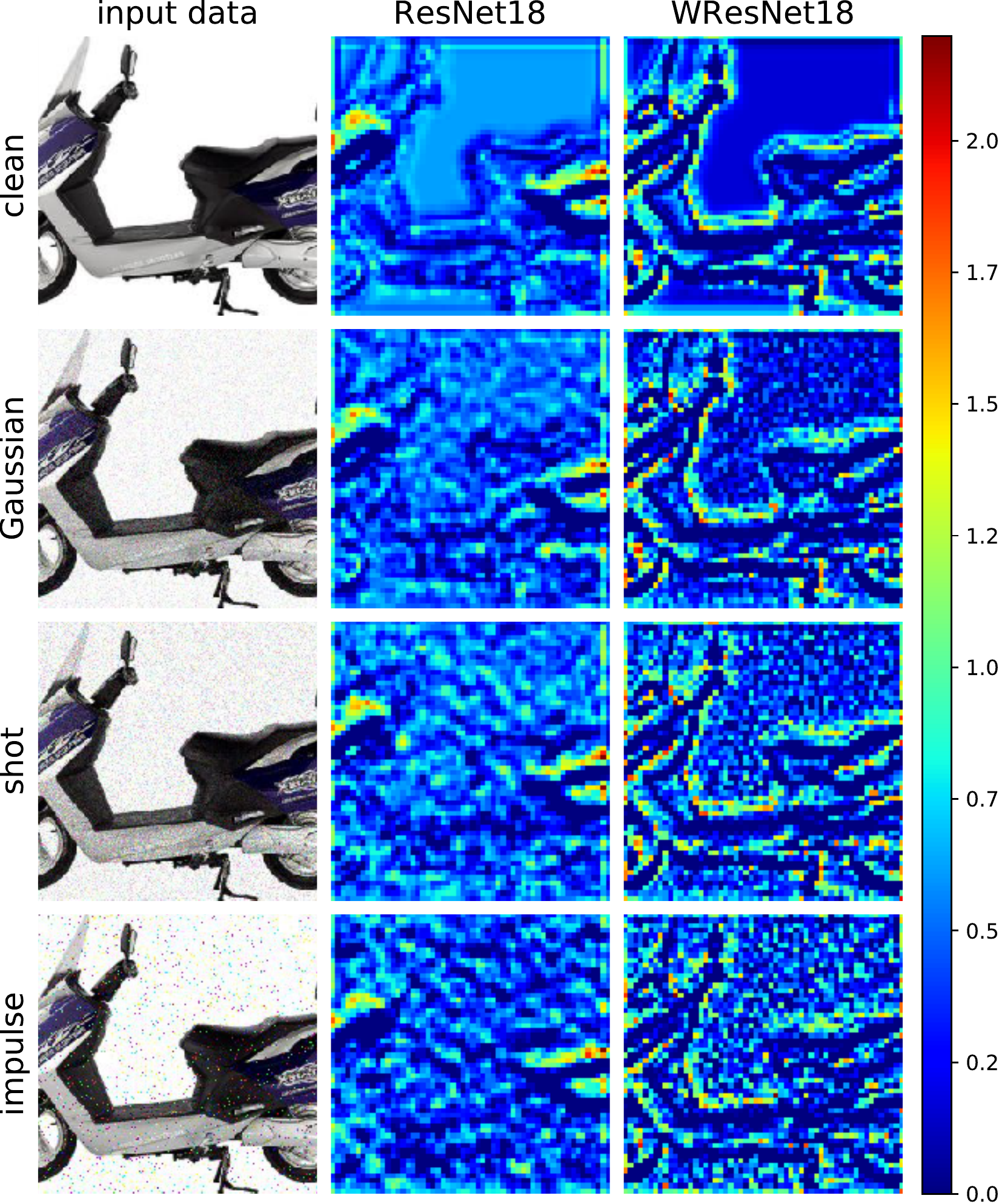}
	\label{fig_feature_map_noise_d}}
	%\subfigure[\emph{schooner}]
	%{\includegraphics*[scale=0.45, viewport=182 68 657 640]{figures/feature_map_noise_new/n04147183_00038543_ResNet18.pdf}
	%\label{fig_feature_map_noise_b}}\\
	%\subfigure[\emph{recreational vehicle}]
	%{\includegraphics*[scale=0.45, viewport=182 68 657 640]{figures/feature_map_noise_new/n04065272_00037537_ResNet18.pdf}
	%\label{fig_feature_map_noise_c}}\hspace{20pt}
	\caption{The feature maps sourced from clean and noisy images.
    {\color{black}In each subfigure, the first column shows the clean example image from ImageNet validation set
    and its three noisy versions corrupted by Gaussian noise, shot noise, and impulse noise, respectively.
    The second column shows their feature maps generated by ResNet18,
    while the third column shows the feature maps generated by WResNet18 using wavelet ``ch2.2''.
    It is difficult for the original CNN to suppress noise,
    while WaveCNet could suppress the noise and maintain the object structure during its inference.}
    }
	\label{fig_feature_map_noise}
\end{figure*}
We test the top-1 errors of WaveCNets and AlexNet on each noise corruption type $c$ at each level of severity $s$,
when WaveCNets and AlexNet are trained on the clean ImageNet training set.
Then, we compute $\text{mCE}_{\text{noise}}^{\text{WaveCNet}}$ according to Eqs. (\ref{eq_CE}) and (\ref{eq_mCE_noise}).
The Table \ref{Tab_mCE_noise_different_versions} shows the detailed results (the ``noisy'' columns).
In Fig. \ref{fig_mCE_noise}, we show the noise mCEs of WaveCNets for different network architectures and various wavelets.
The ``baseline'' corresponds to the noise mCEs of original CNN architectures,
while ``db$p$'', ``ch$p.\tilde{p}$'' and ``haar'' correspond to the mCEs of WaveCNets with different wavelets.
Except VGG16bn, our method obviously increase the noise-robustness of the CNN architectures for image classification.
For example, the noise mCE of ResNet18 (with navy blue color and down triangle marker in Fig. \ref{fig_mCE_noise})
decreases from $88.97$ (``baseline'') to $80.38$ (``ch2.2'').
One can find that all wavelets including ``db5'' and ``db6'' improve the noise-robustness of ResNet18, ResNet34, and ResNet50,
although the classification accuracy of the WResNets with ``db5'' and ``db6'' for the clean images may be lower than that of the original ResNets.
It means that our methods indeed increase the noise-robustness of these network architectures.

Fig. \ref{fig_feature_map_noise} shows two example feature maps for well trained ResNet18 and WResNet18
with clean and noisy images as input.
In each subfigure, the first column shows the clean example image with size of $224\times224$ from ImageNet validation set
and its three noisy versions corrupted by Gaussian noise, shot noise, and impulse noise.
The second column shows their feature maps generated by ResNet18, while the third column shows the feature maps generated by WResNet18 integrated with wavelet ``ch2.2''.
These feature maps are captured from the 16th output channel of the last layer in the network blocks with tensor size of $56\times56$.
From these examples, one can find that it is difficult for the original CNN to suppress noise,
while WaveCNet could suppress the noise and maintain the object structure during its inference.
For example, in Fig. \ref{fig_feature_map_noise_a},
the \emph{vase} structures in the two feature maps generated by both ResNet18 and WResNet18(ch2.2) are complete,
when the clean \emph{vase} image is fed into the networks.
However, after the image is corrupted by noise,
the ResNet18 feature maps contain very strong noise and the \emph{vase} structures vanish,
while the basic structures could still be observed from the WResNet18 feature maps.
For the \emph{moped} in Fig. \ref{fig_feature_map_noise_d},
the network without wavelet, ResNet18, tends to focus its response on the two indigo areas of the \emph{moped} in the feature maps,
and the noise totally breaks the \emph{moped} structures in the feature maps for the three noisy images.
In contrary, wavelet suppresses the noise and the basic \emph{moped} structures are still complete in the feature maps generated by WResNet18,
whether the input image is corrupted by noise or not.
This advantage improves the robustness of WaveCNets against the noise.

The noise-robustness of VGG16bn is inferior to that of ResNet34,
although they achieve similar accuracy ($73.37\%$ and $73.30\%$).
Our method does not significantly improve the noise-robustness of VGG16bn,
although it increases the accuracy by $1.03\%$.
It means that the VGG16bn may not be a proper architecture in terms of noise-robustness.

\subsection{Wavelet denoising block}\label{subsec_wavelet_denoising_block}
\begin{table*}[!t]
	%\scriptsize
	\caption{Noise mCE of WaveCNets with different versions of input data (lower is better).
    {\color{black}Wavelets could improve the noise-robustness of vanilla CNNs.
    The wavelet denoising block consistently increase the noise-robustness of the all CNNs, no matter they are integrated with DWT layer, or not.}}
	\label{Tab_mCE_noise_different_versions}
	\begin{center}
	\begin{threeparttable}
	\setlength{\tabcolsep}{0.888mm}{
	\begin{tabular}{cc||cc||cc||cc||cc||cc||cc}\hline
	\multicolumn{2}{c||}{\multirow{2}{*}{Wavelet}}&\multicolumn{2}{c||}{WVGG16bn} &\multicolumn{2}{c||}{WResNet18}
	&\multicolumn{2}{c||}{WResNet34}&\multicolumn{2}{c||}{WResNet50}&\multicolumn{2}{c||}{WResNet101}&\multicolumn{2}{c}{WDenseNet121} \\\cline{3-14}
	&&noisy\tnote{b}& denoised\tnote{c}&noisy & denoised&noisy & denoised&noisy & denoised&noisy & denoised&noisy & denoised\\\hline\hline
\multicolumn{2}{c||}{none (baseline)\tnote{a}}&               87.62&     86.76{\scriptsize~($-$0.86)}&     88.97&    87.77{\scriptsize~($-$1.20)}&     83.10&    81.52{\scriptsize~($-$1.58)}&
81.31&    79.45{\scriptsize~($-$1.86)}&     75.28&    73.72{\scriptsize~($-$1.56)}&     82.29&  80.92{\scriptsize~($-$1.37)}\\\hline
\multicolumn{2}{c||}{Haar}                    &               87.43&     85.93{\scriptsize~($-$1.50)}&     80.91&    79.68{\scriptsize~($-$1.23)}&     76.63&    75.62{\scriptsize~($-$1.01)}&
78.48&    77.18{\scriptsize~($-$1.30)}&     70.46&    68.50{\scriptsize~($-$1.96)}&     78.33&  77.02{\scriptsize~($-$1.31)}\\\hline
{\multirow{4}{*}{Cohen}}&\multicolumn{1}{|c||}{ch2.2}    &    87.79&     86.18{\scriptsize~($-$1.61)}&\textbf{80.38}& \textbf{78.93}{\scriptsize~($-$1.45)}&     77.61&    75.98{\scriptsize~($-$1.63)}&
76.87&    75.84{\scriptsize~($-$1.03)}&     \textbf{65.06}&    \textbf{64.01}{\scriptsize~($-$1.05)}&     76.97&  76.19{\scriptsize~($-$0.78)}\\
&\multicolumn{1}{|c||}{ch3.3}                 &               88.28&     87.58{\scriptsize~($-$0.70)}&     81.02&    79.80{\scriptsize~($-$1.22)}&     74.30&    73.26{\scriptsize~($-$1.04)}&
75.04&    73.42{\scriptsize~($-$1.62)}&     69.82&    68.46{\scriptsize~($-$1.36)}&     78.31&  76.88{\scriptsize~($-$1.43)}\\
&\multicolumn{1}{|c||}{ch4.4}                 &      \textbf{85.63}&     \textbf{84.67}{\scriptsize~($-$0.96)}&     82.19&    80.66{\scriptsize~($-$1.53)}&     76.19&    74.70{\scriptsize~($-$1.49)}&
76.73&    75.05{\scriptsize~($-$1.68)}&     68.91&    67.62{\scriptsize~($-$1.29)}&     \textbf{76.01}&  \textbf{74.84}{\scriptsize~($-$1.17)}\\
&\multicolumn{1}{|c||}{ch5.5}                 &               89.84&     88.70{\scriptsize~($-$1.14)}&     83.77&    81.97{\scriptsize~($-$1.80)}&     76.00&    74.09{\scriptsize~($-$1.91)}&
\textbf{72.08}&    \textbf{70.99}{\scriptsize~($-$1.09)}&     70.30&    68.90{\scriptsize~($-$1.40)}&     76.27&  75.18{\scriptsize~($-$1.09)}\\\hline
{\multirow{5}{*}{Daubechies}}&\multicolumn{1}{|c||}{db2} &    87.44&     85.97{\scriptsize~($-$1.47)}&     82.54&81.23{\scriptsize~($-$1.31)}&\textbf{72.73}&    \textbf{71.45}{\scriptsize~($-$1.28)}&
78.80&    77.21{\scriptsize~($-$1.59)}&     70.76&    68.82{\scriptsize~($-$1.94)}&     79.82&  78.69{\scriptsize~($-$1.13)}\\
&\multicolumn{1}{|c||}{db3}	                  &88.53    &86.92{\scriptsize~($-$1.61)}&     83.92&    82.02{\scriptsize~($-$1.90)}&     76.43&    74.83{\scriptsize~($-$1.60)}&
79.34&    77.31{\scriptsize~($-$2.03)}&72.70     & 70.97{\scriptsize~($-$1.73)} &79.03    & 78.02{\scriptsize~($-$1.01)}\\
&\multicolumn{1}{|c||}{db4}	                  &90.02               &88.53{\scriptsize~($-$1.49)}&     86.23&    84.32{\scriptsize~($-$1.91)}&     78.96&    77.39{\scriptsize~($-$1.57)}&
77.99&    76.85{\scriptsize~($-$1.14)}&73.50     & 71.89{\scriptsize~($-$1.61)}   &80.89     & 80.12{\scriptsize~($-$0.77)}\\
&\multicolumn{1}{|c||}{db5}	                  &91.26               &89.88{\scriptsize~($-$1.38)}&     85.47&    83.92{\scriptsize~($-$1.55)}&     73.15&    72.03{\scriptsize~($-$1.12)}&
78.33&    77.16{\scriptsize~($-$1.17)}&75.29     & 73.73{\scriptsize~($-$1.56)}   &82.35     & 81.63{\scriptsize~($-$0.72)}\\
&\multicolumn{1}{|c||}{db6}	                  &93.73               &92.12{\scriptsize~($-$1.61)} &     86.57&    84.76{\scriptsize~($-$1.81)}&     82.02&    80.06{\scriptsize~($-$1.96)}&
79.78&    78.61{\scriptsize~($-$1.17)}&78.43     & 76.64{\scriptsize~($-$1.79)}   &84.44     & 83.87{\scriptsize~($-$0.57)}\\\hline
			\end{tabular}}
		\begin{tablenotes}
			\item[a] corresponding to the results of original CNNs, i.e., VGG16bn, ResNets, DenseNet121.
			\item[b] taking the original noisy images in ImageNet-C as input.
			\item[c] taking the denoised version of noisy images in ImageNet-C as input.
		\end{tablenotes}
	\end{threeparttable}
	\end{center}
\end{table*}
While wavelet integrated CNNs like WVGG16bn, WResNets and WDenseNet121
have been show to achieve better classification accuracy on ImageNet and noise-robustness on ImageNet-C,
the wavelet denoising block shown in \ref{fig_denoise_a} can actually be applied to preprocess the noisy images and further improve the performance of various CNNs.

For the noisy images in ImageNet-C, We test a so called soft threshold filtering operation to filter the noises in high-frequency components
$\textbf{\emph{X}}_{lh}, \textbf{\emph{X}}_{hl}, \textbf{\emph{X}}_{hh}$,
and test the performance of different CNNs
when such a denoising is applied, or not.
The commonly used soft threshold filter operation is defined as below:
\begin{align}
\label{eq_hard_threshold}
\text{SoftShrinkage}(x) = \begin{cases}
        x - \lambda, & \text{ if } x > \lambda, \\
        x + \lambda, & \text{ if } x < -\lambda, \\
        0, & \text{otherwise}.
        \end{cases}
\end{align}
For multiple levels of DWT and IDWT, various thresholds $\lambda$, or functions $\lambda(x)$, could be selected.
In our work, we perform one level DWT and IDWT and the value of $\lambda$ is set as 0.1.

Table \ref{Tab_mCE_noise_different_versions} shows the performance of different CNNs
when the wavelet based denoising block is applied to denoise the input images, or not.
The numbers in brackets denote the differences of mCE when the denoising is applied, or not.
One can observe from the table that the denoising module consistently decrease the mCE
for all of the networks like VGG16bn, ResNets and DenseNet121, no matter the CNNs are integrated with our DWT layer, or not.
For vanilla CNNs, the denoising operation reduces the mCE of ResNet50 from 81.31 to 79.45.
For wavelet integrated CNNs, such operation reduces the mCE of WResNet101 from 70.46 to 68.50, when Haar wavelet is concerned.
The lowest mCE (64.01) is achieved by WResNet101 integrated with wavelet Cohen(2,2).
The results also suggest that our proposed DWT integration in CNNs produces a much better improvement of the noise-robustness
than the wavelet based denoising block, e.g., the mCE (80.38) of WResNet18(ch2.2)
using the original noisy input is significantly lower than that of ResNet18 (87.77) input with the denoised data.

\subsection{Comparison with other wavelet based down-sampling}
\begin{figure}[bpt]
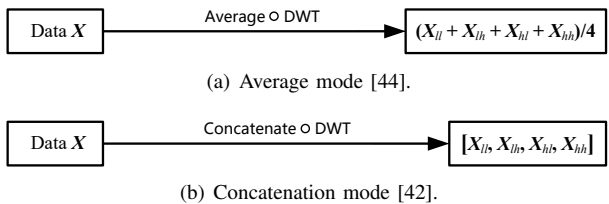

	\centering
	\subfigure[Average mode \cite{duan2017sar}.]{
		\label{fig_wavelet_integrated_modes_b}
		\includegraphics*[scale=0.725, viewport=267 130 585 155]{figures/Visio-down_sampling_of_dwt.pdf}
	}\\
	\subfigure[Concatenation mode \cite{liu2018multi}.]{
		\label{fig_wavelet_integrated_modes_a}
		\includegraphics*[scale=0.725, viewport=267 179 585 204]{figures/Visio-down_sampling_of_dwt.pdf}
	}
	\caption{Wavelet integrated down-sampling in various modes.
    {\color{black}In these two modes, the high-frequency components are added into or concatenated with the low-frequency component,
    which would damage the useful information because of the high-frequency noise.}}
	\label{fig_wavelet_integrated_modes}
\end{figure}
\begin{table}
%\scriptsize
	\caption{Comparison with other wavelet based down-sampling.
    {\color{black}The networks using wavelet based down-sampling generally achieve better accuracy and noise mCE.
    Comprehensively, WResNet18 using DWT$_{ll}$ performs the best among them.
    %The number of parameters of WResNet18\_A is the same with that of original ResNet18, which is similar to WResNet18.
    %However, the added high-frequency noise damage the useful information in the low-frequency component,
    %and WResNet18\_A performs the worst among the networks using wavelet based down-sampling.
    %WResNet18\_C employs much more parameters than WResNet18 and WResNet18\_A, which is almost the same with that for WResNet34.
    %However, its classification performance is much inferior to that of WResNet34.
    }}
	\label{Tab_result_other_modes}
\begin{center}
\setlength{\tabcolsep}{1.00mm}{
\begin{tabular}{l|cccccc|c}\hline
\multicolumn{1}{c|}{\multirow{2}{*}{Network}} &\multicolumn{6}{c|}{Top-1 Accuracy (higher is better)} & \multirow{2}{*}{Params.}\\\cline{2-7}
						 					&haar	&ch2.2	&ch3.3	&ch4.4	&ch5.5	&db2 	&\\\hline
					     WResNet18			&71.47	&71.62		&71.55		&71.52		&71.26		&71.48	&\textbf{11.69}M\\
		WResNet18\_A \cite{duan2017sar} 	&70.06	&69.24		&69.91		&69.98		&70.31		&70.52	&11.69M\\
		WResNet18\_C \cite{liu2018multi} 	&\textbf{71.94}	&\textbf{71.75}		&\textbf{71.66}		&\textbf{71.99}		&\textbf{72.03}		
						 					&\textbf{71.88}	&21.62M\\\cdashline{1-8}[2pt/2pt]
						 WResNet34			&74.35	&74.33		&74.51		&74.61		&74.34		&74.30	&21.80M\\\hline\hline
						 \multirow{2}{*}{ } &\multicolumn{6}{c|}{Noise mCE (lower is better)} & \\\hline
						 %&haar	&ch2.2	&ch3.3	&ch4.4	&ch5.5	&db2 	&\\
						 WResNet18 			&\textbf{80.91}	&\textbf{80.38}		&\textbf{81.02}		&82.19		&83.77		&82.54	& \\
		WResNet18\_A \cite{duan2017sar} 	&83.17	&86.02		&86.07		&85.22		&82.96		&84.01	& \\
		WResNet18\_C \cite{liu2018multi} 	&81.79	&83.67		&83.51		&\textbf{82.13}		&\textbf{82.60}		
						 					&\textbf{80.11}	& \\\cdashline{1-8}[2pt/2pt]
						 WResNet34			&76.64	&77.61		&74.30		&76.19		&76.00		&72.73	& \\\hline
\end{tabular}}
\end{center}
\end{table}
Different with our wavelet based down-sampling (Fig. \ref{fig_denoise_b}),
there are other wavelet integrated down-sampling modes in literatures.
In \cite{duan2017sar}, the authors adopt as down-sampling output the average value of the multiple components of wavelet transform,
as Fig. \ref{fig_wavelet_integrated_modes_b} shows.
In \cite{liu2018multi}, the authors concatenate all the components output from DWT, and process them in a unified way,
as Fig. \ref{fig_wavelet_integrated_modes_a} shows.

Taking ResNet18 as backbone, we compare our wavelet based down-sampling with the previous approaches,
in terms of classification accuracy and noise-robustness.
%although they are used for different tasks in previous literatures.
We rebuild ResNet18 using the three down-sampling modes shown in Fig. \ref{fig_denoise_b} and Fig. \ref{fig_wavelet_integrated_modes},
and denote them as WResNet18, WResNet18\_A, and WResNet18\_C, respectively.
We train them on ImageNet when various wavelets are used.
Table \ref{Tab_result_other_modes} shows the accuracy on ImageNet
and the noise mCEs on the ImageNet-C.
Generally, the networks using wavelet based down-sampling achieve better accuracy and noise mCE
than that of original ResNet18 ($69.76\%$ accuracy and $88.97$ mCE).

Similar to WResNet18, the number of parameters of WResNet18\_A is the same with that of original ResNet18.
However, the added high-frequency components in the feature maps damage the information contained in the low-frequency component,
because of the high-frequency noise.
WResNet18\_A performs the worst among the networks using wavelet based down-sampling.
\begin{table*}[!tbp]
	%\scriptsize
    \centering
	%\small
	\caption{Accuracy of WaveCNets on adversarial samples.
    {\color{black}The adversarial noises significantly decrease the classification accuracy of well-trained CNNs,
    while the wavelets could increase the performance of CNNs on the various adversarial samples.
    %Take PGD attack for example, our WResNet101 using wavelet Haar, Cohen, and Daubechies with short filters achieve about 3\% higher accuracy.
    %The best performance is achieved by ``ch2.2'', which is abut 4.35\% higher than that of ResNet101.
    }}
	\label{Tab_WaveCNet_accuracy_ad}
	\begin{threeparttable}
		\setlength{\tabcolsep}{2.25mm}
			\begin{tabular}{cc||cccccc}\hline
\multicolumn{2}{c||}{\multirow{2}{*}{Network}}&\multicolumn{6}{c}{Attacker}\\\cline{3-8}
        &&FGSM &IFGSM&iterLL&RFGSM&CW&PGD\\\hline\hline
{\multirow{11}{*}{WResNet101}}
&\multicolumn{1}{|c||}{none\tnote{*}} &69.43         &66.53         &72.25         &50.52         &68.76         &53.94 \\\cline{2-8}
&\multicolumn{1}{|c||}{haar} &70.91~($+$1.47)&68.33~($+$1.80)&73.32~($+$1.08)&53.22~($+$2.70)&70.03~($+$1.27)&57.30~($+$3.36)\\\cdashline{2-8}[2pt/2pt]
&\multicolumn{1}{|c||}{ch2.2}&\textbf{71.81~($+$2.38)}&\textbf{69.22~($+$2.69)}&\textbf{74.33~($+$2.09)}&\textbf{54.08~($+$3.56)}&70.82~($+$2.06)&\textbf{58.28~($+$4.35)}\\
&\multicolumn{1}{|c||}{ch3.3}&71.55~($+$2.12)&68.83~($+$2.29)&74.22~($+$1.97)&52.86~($+$2.34)&70.86~($+$2.10)&56.99~($+$3.06)\\
&\multicolumn{1}{|c||}{ch4.4}&71.32~($+$1.89)&68.58~($+$2.05)&74.02~($+$1.77)&53.15~($+$2.63)&70.45~($+$1.69)&56.84~($+$2.90)\\
&\multicolumn{1}{|c||}{ch5.5}&71.31~($+$1.88)&68.61~($+$2.07)&74.03~($+$1.78)&52.62~($+$2.10)&\textbf{70.87~($+$2.11)}&56.93~($+$3.00)\\\cdashline{2-8}[2pt/2pt]
&\multicolumn{1}{|c||}{db2}  &71.66~($+$2.23)&68.81~($+$2.28)&74.25~($+$2.00)&52.56~($+$2.04)&70.75~($+$1.99)&56.80~($+$2.86)\\
&\multicolumn{1}{|c||}{db3}  &70.47~($+$1.04)&67.71~($+$1.18)&73.06~($+$0.81)&51.67~($+$1.15)&69.22~($+$0.46)&55.69~($+$1.75)\\
&\multicolumn{1}{|c||}{db4}  &69.56~($+$0.13)&66.89~($+$0.36)&72.12~($-$0.13)&51.10~($+$0.58)&68.11~($-$0.65)&55.17~($+$1.23)\\
&\multicolumn{1}{|c||}{db5}  &68.88~($-$0.55)&66.21~($-$0.32)&71.08~($-$1.17)&50.16~($-$0.36)&67.16~($-$1.60)&54.41~($+$0.47)\\
&\multicolumn{1}{|c||}{db6}  &67.85~($-$1.58)&65.12~($-$1.41)&69.98~($-$2.27)&49.45~($-$1.07)&65.74~($-$3.02)&53.74~($-$0.20)\\\hline\hline
{\multirow{11}{*}{WDenseNet121}}
&\multicolumn{1}{|c||}{none} &65.92         &62.46         &68.38         &46.11         &64.66         &48.74 \\\cline{2-8}
&\multicolumn{1}{|c||}{haar} &67.27~($+$1.35)&63.92~($+$1.46)&69.79~($+$1.41)&47.92~($+$1.81)&66.35~($+$1.69)&51.58~($+$2.84)\\\cdashline{2-8}[2pt/2pt]
&\multicolumn{1}{|c||}{ch2.2}&\textbf{67.81~($+$1.89)}&64.59~($+$2.13)&70.15~($+$1.77)&\textbf{48.91~($+$2.80)}&66.13~($+$1.47)&52.43~($+$3.69)\\
&\multicolumn{1}{|c||}{ch3.3}&67.62~($+$1.70)&64.45~($+$1.99)&70.13~($+$1.75)&48.35~($+$2.34)&66.31~($+$1.65)&51.83~($+$3.09)\\
&\multicolumn{1}{|c||}{ch4.4}&67.75~($+$1.84)&\textbf{64.79~($+$2.33)}&\textbf{70.18~($+$1.80)}&48.83~($+$2.72)&66.47~($+$1.81)&\textbf{52.53~($+$3.78)}\\
&\multicolumn{1}{|c||}{ch5.5}&67.71~($+$1.79)&64.60~($+$2.14)&70.11~($+$1.73)&48.43~($+$2.31)&66.72~($+$2.06)&51.96~($+$3.22)\\\cdashline{2-8}[2pt/2pt]
&\multicolumn{1}{|c||}{db2}  &67.39~($+$1.47)&64.17~($+$1.71)&69.98~($+$1.60)&47.64~($+$1.53)&\textbf{66.73~($+$2.07)}&51.11~($+$2.37)\\
&\multicolumn{1}{|c||}{db3}  &67.18~($+$1.26)&63.81~($+$1.35)&69.65~($+$1.27)&47.54~($+$1.43)&65.83~($+$1.17)&51.02~($+$2.28)\\
&\multicolumn{1}{|c||}{db4}  &66.02~($+$0.10)&62.80~($+$0.34)&68.27~($-$0.11)&46.60~($+$0.49)&64.49~($-$0.17)&50.19~($+$1.45)\\
&\multicolumn{1}{|c||}{db5}  &64.85~($-$1.07)&61.63~($-$0.83)&66.81~($-$1.57)&46.06~($-$0.05)&62.95~($-$1.71)&49.42~($+$0.68)\\
&\multicolumn{1}{|c||}{db6}  &63.47~($-$2.45)&60.30~($-$2.16)&65.16~($-$3.22)&44.75~($-$1.36)&61.18~($-$3.48)&48.56~($-$0.18)\\\hline
			\end{tabular}
		\begin{tablenotes}
			%\tiny
			\item[*] corresponding to the accuracy of original CNNs on the adversarial samples, i.e., ResNet101, DenseNet121.
		\end{tablenotes}
	\end{threeparttable}
\end{table*}

Due to the tensor concatenation,
WResNet18\_C employs much more parameters ($21.62 \times 10^6$) than WResNet18 and WResNet18\_A ($11.69 \times 10^6$).
WResNet18\_C thus increase the accuracy of WResNet18 by $0.11\%$ to $0.77\%$, when various wavelets are used.
However, due to the included noise,
the concatenation does not evidently improve the noise-robustness.
In addition, the amount of parameters for WResNet18\_C is almost the same with that for WResNet34 ($21.80\times10^6$),
while the accuracy and noise mCE of WResNet34 are obviously superior to that of WResNet18\_C.

\subsection{Adversarial robustness}\label{subsec_adversarial_robustness}
A lot of researches \cite{goodfellow2014explaining,Kurakin2017adversarial,Tramer2018ensemble,Carlini2016towards,Madry2016towards,xie2019feature}
have shown that CNNs can be fooled by adversarial samples, which are generated by adding specially designed noises to the normal samples.
While such noises are usually high-frequency components \cite{wang2020high} and not noticeable to human eyes,
we believe that our WaveCNets can filter such noises and increase the robustness of the CNNs against such adversarial attacks.
%Well-trained classifier could be fooled by adversarial examples, which are only modified slightly with elaborate attack algorithms
%\cite{goodfellow2014explaining,Kurakin2017adversarial,Tramer2018ensemble,Carlini2016towards,Madry2016towards}.
%The modification are usual high-frequency components in the adversarial images that human cannot notice.
%Therefore, the adversarial examples could be denoised by wavelet transforms,
%and WaveCNets would achieve higher adversarial robustness than the original CNNs.

We test the adversarial robustness of WaveCNets using six attack algorithms,
i.e., fast gradient sign method (FGSM) \cite{goodfellow2014explaining},
iterative FGSM (IFGSM) \cite{Kurakin2017adversarial},
iterative least-likerly class attack (iterLL) \cite{Kurakin2017adversarial},
random FGSM (RFGSM) \cite{Tramer2018ensemble},
Carlini and Wagner's $L_2$ attacker (CW) \cite{Carlini2016towards},
and projected gradient descent (PGD) \cite{Madry2016towards}.
Using a public code of the above attackers\footnote{\href{/https://github.com/Harry24k/adversarial-attacks-pytorch}{/https://github.com/Harry24k/adversarial-attacks-pytorch}},
we generate adversarial samples using the ImageNet validation set with the scenario of black-box attacks,
and the attack model is ``InceptionV3'' trained on ImageNet.

Table \ref{Tab_WaveCNet_accuracy_ad} shows the accuracy of WaveCNets (WResNet101 and WDenseNet121) on the six adversarial sample sets;
the classifiers are only trained on the clean ImageNet training set.
In Table \ref{Tab_WaveCNet_accuracy_ad}, ``none'' corresponds to the accuracy of original CNNs (ResNet101 and DenseNet121)
on ImageNet validation set attracted by the six attack algorithms.
The parenthesized numbers are accuracy differences of WaveCNets compared with the CNN results.
From Table \ref{Tab_WaveCNet_accuracy_ad},
we find that the wavelets could increase the accuracy of CNNs on the adversarial samples.
Take PGD attack for example, our WResNet101 using different wavelets generally achieve 3\% higher accuracy than the original ResNet101 (53.94\%).
The best performance is achieved by ``ch2.2'', which is abut 4.35\% higher than that of ResNet101.
%For example, attacked by PGD, the accuracy of ResNet101 on the ImageNet validation set decreases from $77.37\%$ to $53.94\%$,
%while ResNet101 integrated with wavelet ``ch2.2'' achieves the accuracy of $58.28\%$, with an accuracy increase of $4.35\%$.

{\color{black}
While our WaveCNets achieve consistent resistance to various attackers,
their performances are inferior to that of the specially trained defense methods \cite{xie2019feature}.
With adversarial training, the feature denoising method proposed in \cite{xie2019feature} achieves the state-of-the-art result in defending adversarial attackers.
However, as the authors point out in their paper, the custom-designed denoising block requires residual connection for stable training of their deep networks.
It seems that the spatial filtering used in their method, such as mean filtering and median filtering, tend to perform denoising in the whole frequency domain,
which might easily break basic object structures in the feature maps.
In contrast, the general DWT module in WaveCNets denoises the feature maps in the high-frequency interval,
which can keep the basic object structure
and thus lead to increased adversarial-robustness without requirement of adversarial training.
}
%Wavelet efficiently defends the CNNs against various attacks.
\subsection{Object detection}

{\color{black}To further illustrate the efficiency of wavelets in deep learning,
we conduct detection experiments on COCO \cite{COCO_2014_ECCV} benchmark,
by comparing the performance of faster R-CNN \cite{faster_rcnn}, RetinaNet \cite{Lin_2017_RetinaNet}, and their wavelet integrated versions.
We apply the original off-the-shelf detectors in mmdetection \cite{mmdetection},
and build their wavelet integrated versions by replacing original backbones (ResNet50 and ResNet101) with WaveCNets (WResNet50 and WResNet101).
%applying WaveCNet (i.e., WResNet50 and WResNet101) based backbones instead of original CNNs (ResNet50 and ResNet101).
In the wavelet integrated detectors,
the DWT layer is applied to down-sample the feature maps and suppress the aliasing effects.
Wavelet ``haar'' and ``ch3.3'' are applied into these two backbones,
since they show the best performance on ImageNet classification task.}
{\color{black}}
\begin{table*}[!tbp]
    %\scriptsize
	%\small
    \caption{
    {\color{black}Results of wavelet integrated detectors on COCO.
            The detection performance of original detectors are taken as baseline.
            For all detectors, our method could improve their detection performance.
            Wavelet consistently increase the detectors' performance on all three categories of objects, i.e., small, medium, and large objects.}}
	\label{Tab_AP_COCO}
	\begin{center}
	\begin{threeparttable}
	\setlength{\tabcolsep}{1.mm}{
	\begin{tabular}{c|c||c|c|c||c|ccc|ccc|ccc|ccc}\hline
    \multicolumn{1}{c|}{}      &\multirow{2}{*}{Backbone}& \multirow{2}{*}{initial lr}&\multirow{2}{*}{BS\tnote{a}}& \multirow{2}{*}{Epoch} & \multirow{2}{*}{FPS} & \multicolumn{6}{c|}{\texttt{minival}}& \multicolumn{6}{c}{\texttt{test-dev}}\\\cline{7-18}
                               &&&&&&AP &  AP$_{50}$ & AP$_{75}$ & AP$_{S}$ & AP$_{M}$ & AP$_{L}$&AP &  AP$_{50}$ & AP$_{75}$ & AP$_{S}$ & AP$_{M}$ & AP$_{L}$\\\hline
    \multirow{14}{*}{faster R-CNN}   &ResNet50  & 0.02 & 16& 12    & \textbf{18.1}
                     &$37.3$&$58.2$&$40.4$&$21.4$&$41.0$&$48.5$
                        &$37.7$&$58.7$&$41.0$&$21.8$&$40.6$&$46.8$\\
   &WResNet50(haar)  &  0.02 & 16& 12     &16.3
                     &$\textbf{38.5}$&$\textbf{59.8}$&$\textbf{41.7}$&$\textbf{22.6}$&$\textbf{42.4}$&$\textbf{49.2}$
                        &$\textbf{38.7}$&$\textbf{60.2}$&$\textbf{42.1}$&$\textbf{22.7}$&$\textbf{41.7}$&$\textbf{47.8}$\\
      &&&&&$^{-1.8}$&$^{+1.2}$&$^{+1.6}$&$^{+1.3}$&$^{+1.2}$&$^{+1.4}$&$^{+0.7}$
                        &$^{+1.0}$&$^{+1.5}$&$^{+1.1}$&$^{+0.9}$&$^{+1.1}$&$^{+1.0}$\\\cdashline{2-18}[2pt/2pt]
                         &ResNet50  & 0.02 & 16&24\tnote{b}& \textbf{18.1}
                     &$38.4$&$59.0$&$42.0$&$21.5$&$42.1$&$50.3$
                        &$38.7$&$59.6$&$42.1$&$22.1$&$41.4$&$48.6$\\
      &&&&&$^{~~~0}$&$^{+1.1}$&$^{+0.8}$&$^{+1.6}$&$^{+0.1}$&$^{+1.1}$&$^{+1.8}$
                        &$^{+1.0}$&$^{+0.9}$&$^{+1.1}$&$^{+0.3}$&$^{+0.8}$&$^{+1.8}$\\
                    &WResNet50(haar)    &0.01&8&24&16.3&$\textbf{39.4}$&$\textbf{60.7}$&$\textbf{42.6}$&$\textbf{22.1}$&$\textbf{43.3}$&$\textbf{50.3}$
                    &$\textbf{39.5}$&$\textbf{60.6}$&$\textbf{43.0}$&$\textbf{22.8}$&$\textbf{43.2}$&$\textbf{49.4}$\\
      &&&&&$^{-1.8}$&$^{+2.1}$&$^{+2.5}$&$^{+2.2}$&$^{+0.7}$&$^{+2.3}$&$^{+1.8}$&$^{+1.8}$&$^{+1.9}$&$^{+2.0}$&$^{+1.0}$&$^{+2.6}$&$^{+2.6}$\\\cline{2-18}
                        &ResNet101  & 0.01 & 8 & 12&\textbf{14.8}
                            &$39.5$&$60.1$&$43.3$&$22.3$&$43.5$&$51.5$
                                &$39.8$&$60.9$&$43.4$&$23.0$&$43.0$&$50.0$\\
                        &WResNet101(ch3.3)  &   0.01 & 8 & 12      &12.5  &$\textbf{40.8}$&$\textbf{61.8}$&$\textbf{44.6}$&$\textbf{23.5}$&$\textbf{44.9}$&$\textbf{53.0}$
                            &$\textbf{41.2}$&$\textbf{62.6}$&$\textbf{45.1}$&$\textbf{24.1}$&$\textbf{44.7}$&$\textbf{51.2}$\\
                        &&&&&$^{-2.3}$&$^{+1.3}$&$^{+1.7}$&$^{+1.3}$&$^{+1.2}$&$^{+1.4}$&$^{+1.5}$
                            &$^{+1.4}$&$^{+1.7}$&$^{+1.7}$&$^{+1.1}$&$^{+1.7}$&$^{+1.2}$\\\cdashline{2-18}[2pt/2pt]
                         &ResNet101  & 0.02 & 16&24    & \textbf{14.8}
                     &$39.8$&$60.1$&$43.3$&$22.5$&$43.6$&$52.8$
                        &$40.5$&$61.2$&$44.1$&$22.3$&$43.7$&$51.6$\\
                        &&&&&$^{~~~0}$&$^{+0.3}$&$^{+0.0}$&$^{+0.0}$&$^{+0.2}$&$^{+0.1}$&$^{+1.3}$
                            &$^{+0.7}$&$^{+0.3}$&$^{+0.7}$&$^{-0.7}$&$^{+0.7}$&$^{+1.6}$\\
                    &WResNet101(ch3.3)    &0.01&8&24&12.5&$\textbf{40.9}$&$\textbf{61.5}$&$\textbf{44.6}$&$\textbf{23.1}$&$\textbf{44.9}$&$\textbf{54.1}$
                    &$\textbf{41.5}$&$\textbf{62.4}$&$\textbf{45.1}$&$\textbf{23.5}$&$\textbf{44.6}$&$\textbf{52.5}$\\
                    &&&&&$^{-2.3}$&$^{+1.4}$&$^{+1.4}$&$^{+1.3}$&$^{+0.8}$&$^{+1.4}$&$^{+2.6}$&$^{+1.7}$&$^{+1.5}$&$^{+1.7}$&$^{+0.5}$&$^{+1.6}$&$^{+2.5}$\\\hline
    \multirow{14}{*}{RetinaNet}      &ResNet50   & 0.005 & 8 & 12    &\textbf{17.2}
                    &$36.5$&$55.9$&$39.0$&$21.0$&$40.2$&$46.4$
                        &$36.9$&$56.3$&$39.4$&$20.8$&$39.9$&$45.3$\\
                    &WResNet50(haar)  &  0.005 & 8 & 12     &15.2 &$\textbf{37.4}$&$\textbf{57.0}$&$\textbf{40.2}$&$\textbf{21.9}$&$\textbf{41.4}$&$\textbf{48.0}$
                        &$\textbf{37.7}$&$\textbf{57.6}$&$\textbf{40.4}$&$\textbf{22.0}$&$\textbf{40.6}$&$\textbf{45.9}$\\
                    &&&&&$^{-2.0}$&$^{+0.9}$&$^{+1.1}$&$^{+1.2}$&$^{+0.9}$&$^{+1.2}$&$^{+1.6}$
                        &$^{+0.8}$&$^{+1.3}$&$^{+1.0}$&$^{+1.2}$&$^{+0.7}$&$^{+0.6}$\\\cdashline{2-18}[2pt/2pt]
                         &ResNet50  & 0.01 & 16&24    & \textbf{17.2}
                     &$37.4$&$56.7$&$39.6$&$20.0$&$40.7$&$\textbf{49.7}$
                        &$37.7$&$57.2$&$40.2$&$20.4$&$40.2$&$\textbf{48.1}$\\
                    &&&&&$^{~~~0}$&$^{+0.9}$&$^{+0.8}$&$^{+0.6}$&$^{-1.0}$&$^{+0.5}$&$^{+3.3}$
                        &$^{+0.8}$&$^{+0.9}$&$^{+0.8}$&$^{-0.4}$&$^{+0.3}$&$^{+2.8}$\\
                    &WResNet50(haar)    &0.005&8&24&15.2&$\textbf{38.0}$&$\textbf{57.5}$&$\textbf{40.4}$&$\textbf{21.5}$&$\textbf{41.3}$&$49.4$
                    &$\textbf{38.2}$&$\textbf{58.1}$&$\textbf{40.8}$&$\textbf{21.4}$&$\textbf{40.6}$&$47.6$\\
                    &&&&&$^{-2.0}$&$^{+1.5}$&$^{+1.6}$&$^{+1.4}$&$^{+0.5}$&$^{+1.1}$&$^{+3.0}$&$^{+1.3}$&$^{+1.8}$&$^{+1.4}$&$^{+0.6}$&$^{+0.7}$&$^{+2.3}$\\\cline{2-18}
                    &ResNet101  & 0.005 & 8 &  12    &\textbf{13.6}
                        &$38.5$&$57.9$&$41.4$&$22.1$&$43.2$&$49.7$
                            &$39.1$&$58.7$&$41.9$&$22.4$&$42.3$&$48.7$\\
                        &WResNet101(ch3.3)  &  0.005 & 8 & 12      &11.5
                            &$\textbf{40.1}$&$\textbf{60.1}$&$\textbf{43.1}$&$\textbf{23.7}$&$\textbf{44.2}$&$\textbf{51.9}$
                                &$\textbf{40.2}$&$\textbf{60.6}$&$\textbf{43.1}$&$\textbf{23.4}$&$\textbf{43.7}$&$\textbf{49.7}$\\
                                    &&&&&$^{-2.1}$&$^{+1.6}$&$^{+2.2}$&$^{+1.7}$&$^{+1.6}$&$^{+1.0}$&$^{+2.2}$&$^{+1.1}$&$^{+1.9}$&$^{+1.2}$&$^{+1.0}$&$^{+1.4}$&$^{+1.0}$\\\cdashline{2-18}[2pt/2pt]
                         &ResNet101  & 0.01 & 16&24    & \textbf{13.6}
                     &$38.9$&$58.0$&$41.5$&$21.0$&$42.8$&$52.4$
                        &$39.6$&$59.1$&$42.4$&$21.3$&$42.6$&$\textbf{51.1}$\\
                    &&&&&$^{~~~0}$&$^{+0.3}$&$^{+0.1}$&$^{+0.1}$&$^{-1.1}$&$^{-0.4}$&$^{+2.7}$
                        &$^{+0.5}$&$^{+0.4}$&$^{+0.5}$&$^{-1.1}$&$^{+0.3}$&$^{+2.4}$\\
                    &WResNet101(ch3.3)    &0.005&8&24&11.5
                    &$\textbf{39.9}$&$\textbf{59.6}$&$\textbf{42.8}$&$\textbf{22.4}$&$\textbf{43.7}$&$\textbf{52.7}$
                    &$\textbf{40.1}$&$\textbf{60.1}$&$\textbf{43.1}$&$\textbf{22.4}$&$\textbf{43.1}$&$50.6$\\
                    &&&&&$^{-2.1}$&$^{+1.4}$&$^{+1.7}$&$^{+1.4}$&$^{+0.3}$&$^{+0.5}$&$^{+3.0}$&$^{+1.0}$&$^{+1.4}$&$^{+1.2}$&$^{+0.0}$&$^{+0.8}$&$^{+1.9}$\\\hline
			\end{tabular}}
		\begin{tablenotes}
            \item[a] {\color{black}in mmdetection, the default ratio of initial learning rate to batch size is $\frac{1}{800}$ for faster R-CNN, and $\frac{1}{1600}$ for RetinaNet.}
			\item[b] {\color{black}the original detectors trained for 24 epochs are downloaded from mmdetection. In their training, the learning rates are decayed at epoch 16 and 22.}
		\end{tablenotes}
	\end{threeparttable}
	\end{center}
\end{table*}
\begin{figure*}[bpt]
    \color{black}
	\centering
	\subfigure[Two example images detected by RetinaNet.]
	{\includegraphics*[scale=0.625, viewport=28 310 815 580]{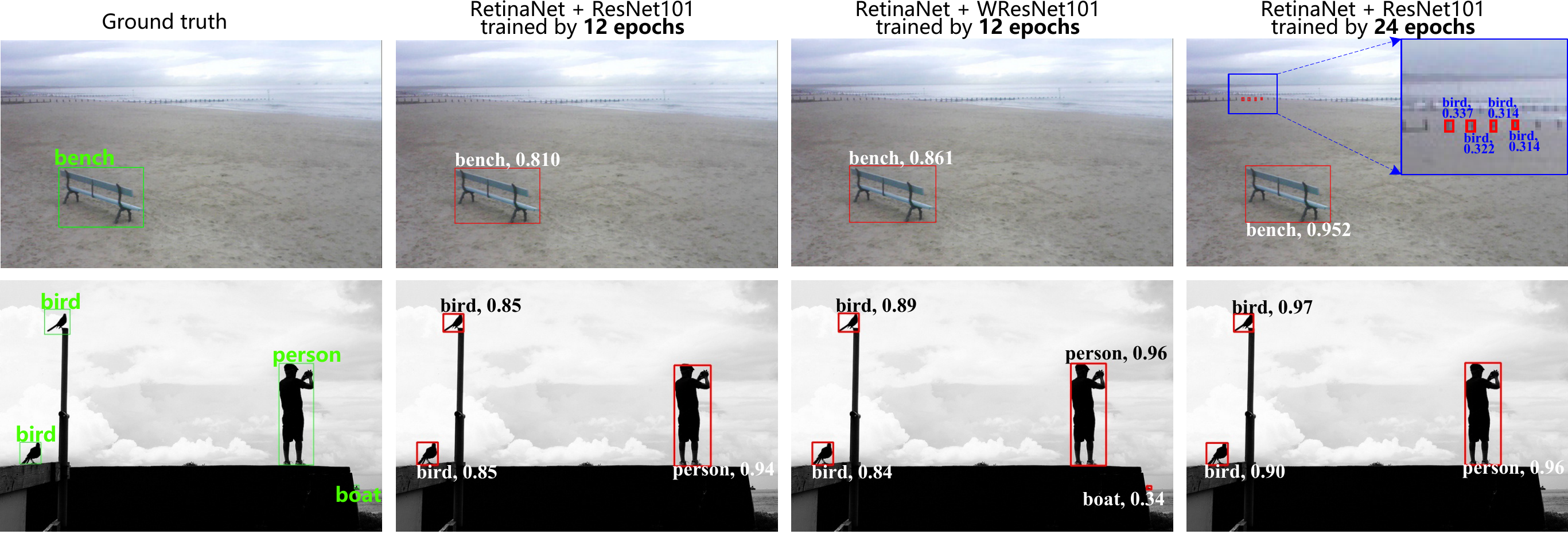}
	\label{fig_detection_retinanet}}
	\subfigure[Two example images detected by Faster R-CNN.]
	{\includegraphics*[scale=0.625, viewport=28 259 815 558]{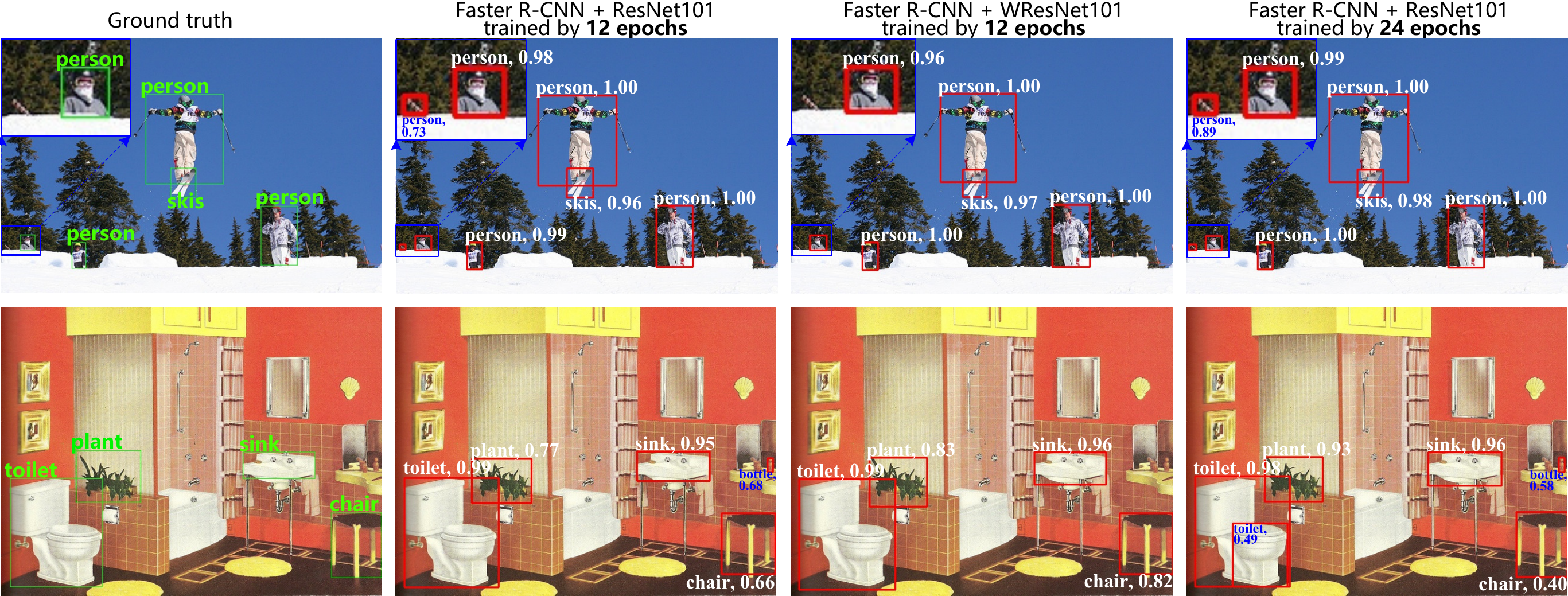}
	\label{fig_detection_faster_rcnn}}
	\caption{\color{black}
    {Four example images detected by RetinaNet and Faster R-CNN, integrated with or without wavelet.
    The manual detections are taken as ground truths.
    While the baseline detectors (trained for 24 epochs) falsely detect or miss the small target objects,
    the detectors integrated with wavelet (trained for only 12 epochs) correctly detect the target objects.
    %Integrated with wavelet ``ch3.3'', the detectors could correctly detect the target objects.
    %Although the baseline detector trained for 24 epochs detects the bench with higher probability ($0.952$),
    %it falsely detects some poles as birds.
    }
    }
	\label{fig_detection_comparison}
\end{figure*}

{\color{black}With the default hyper-parameters, the detectors are trained on \texttt{trainval135k} (115K images) of COCO.
Table \ref{Tab_AP_COCO} shows the hyper-parameter setting.
We train each detector for 12 epochs from scratch, and drop the learning rate by a factor of 0.1 at epoch 8 and 11.
The trained detectors are tested on the \texttt{minival} (5K images) and \texttt{test-dev} (20K images) sets;
Table \ref{Tab_AP_COCO} shows the detection results.
In the experiments, AP, AP$_{50}$, AP$_{75}$,
AP$_S$, AP$_M$, and AP$_L$ are used to evaluate the detection performance,
where
\begin{itemize}
  \item AP: mean value of 10 average precisions (AP) for intersection over union (IoU) larger than 0.50, 0.55, 0.60, $\cdots$, and 0.95;
  \item AP$_{50}$: AP at IoU $\geq 0.50$;
  \item AP$_{75}$: AP at IoU $\geq 0.75$;
  \item AP$_{S}$: AP for small objects (area $\leq 32^2$);
  \item AP$_{M}$: AP for medium objects ($32^2<$ area $\leq96^2$);
  \item AP$_{L}$: AP for large objects ($96^2<$ area).
\end{itemize}
The performances of original detectors trained by 12 epochs are taken as baseline,
and the numbers with smaller size in Table \ref{Tab_AP_COCO} indicate the performance difference with the baseline.
For comparison, we also list the detection performance of original detectors trained by 24 epochs.
}

{\color{black}
As shown in Table \ref{Tab_AP_COCO}, with 12-epoch training,
the detection performance of wavelet integrated detectors are superior to that of the original detectors,
and comparable to or even better than that of original detectors trained by 24 epochs.
Taking faster R-CNN with ResNet101 for example, on \texttt{test-dev} set,
the AP of the 12-epoch trained detector is $39.8\%$, which is increased by $1.4\%$ to $41.2\%$, after integrating with wavelet ``ch3.3''.
The increased AP is even higher than that of original detector ($40.5\%$) trained by 24 epochs.
Comparing their detection performance on the objects with different sizes,
one can find that the wavelet could consistently improve the detector performance on the all three categories of objects.
%The performance on the medium and large objects can further improved by extending the training time,
%while the detection performance on the small objects would be decreased by such extension.
For faster R-CNN with ResNet101 on \texttt{test-dev} set,
the wavelet integrated detector increases the performance on the small, medium, and large objects by $1.1\%, 1.7\%$, and $1.2\%$, respectively.
While extension of training time to 24 epochs could further increase the AP$_M$ and AP$_L$ of the baseline detector by $0.7\%$ and $1.6\%$,
such extension decreases AP$_S$ by $0.7\%$.
Since aliasing effects are available in original detectors, extending training time seems to enlarge such effects.
As the basic structures of small objects are more easily damaged by the aliasing effects,
the detection performance on small objects decreases with the increase of training epochs.
In contrast, wavelet could suppress the aliasing effects in the deep networks,
and consistently increase the performance of detectors for objects with different sizes.
%The above results may be resulted from that the original deep detectors suffer the aliasing effects,
%which could easily damaged the basic structures of small objets,
%and extending training time would enlarge the effects.
%The basic structures of small objects are easily damaged by the aliasing effects.
%Therefore, the detection performance on small objects is decreased as the training time increases.
%Our method could suppress the aliasing effects in the deep networks,
%consistently increasing the performance of detectors on the objects with different sizes.
%making the deep detectors to focus on small and medium objects besides the large objects.
}

{\color{black}
Fig. \ref{fig_detection_comparison} visually shows four example images detected by various detectors,
i.e., Faster R-CNNs and RetinaNets with different backbones (ResNet101 and WResNet101).
In Fig. \ref{fig_detection_comparison}, the first column shows the manual detections, which are taken as ground truths.
While the second and third columns show the images detected by detectors trained for 12 epochs, integrated with or without using wavelet,
the fourth column shows the images detected by the one trained for 24 epochs without using wavelet.
%While Fig. \ref{fig_detection_gt} shows the manual detection,
%Fig. \ref{fig_detection_retinanet_r101_12ep} - \ref{fig_detection_retinanet_r101_24ep} show the results detected by the three RetinaNets.
As Fig. \ref{fig_detection_comparison} shows, the detectors without using wavelet usually falsely detect or miss the small target objects in the images,
while the detectors integrated with wavelet correctly detect all target objects.
Taking the ``bench'' image (shown in the first row of Fig. \ref{fig_detection_comparison}) for example,
the confidence of detected bench increases from $0.810$ to $0.861$ when wavelet ``ch3.3'' is integrated to the backbone of ResNet101.
Though increase of training epoch (24) could also improve such confidence of baseline RetinaNet with backbone of ResNet101, it falsely detects some poles as birds.
%Fig. \ref{fig_detection_gt} shows the manually detection result, which is taken as the ground truth.
%Fig. \ref{fig_detection_retinanet_r101_12ep} - \ref{fig_detection_retinanet_r101_24ep} shows the results detected by 12-epoch trained RetinaNet
%As Fig. \ref{fig_detection_gt} shows, a bench is manually detected in the example image,
%which can also be detected by the three RetinaNets (Fig. \ref{fig_detection_retinanet_r101_12ep} - \ref{fig_detection_retinanet_r101_24ep}).
%With wavelet ``ch3.3'', the 12-epoch trained detector increases the probability for the detected bench from $0.810$ to $0.861$.
%While the 24-epoch trained detector detects the target bench with even higher probability ($0.952$),
%however, it falsely detects some poles as birds, with probabilities higher than $0.3$.
%The above result illustrates the wavelet efficiency in deep network for image detection.
}

{\color{black}
Compared with the commonly used down-sampling, DWT needs more multiply-add operations,
which would decreases the detection speed.
As shown in the 6th column of Table \ref{Tab_AP_COCO}, the detection speed of faster R-CNN slightly decreases from 18.1 FPS (frames per second) to 16.3 FPS,
when wavelet ``haar'' is integrated to the backbone of ResNet50 and Nvidia Tesla V100 GPU is used as the computing platform.
%On a Nvidia Tesla V100 GPU, we test the speed of the various detectors,
%and the results are shown in the 6th column of Table \ref{Tab_AP_COCO}.
%For example, the detection speed of faster R-CNN with ResNet50 decreases from 18.1 FPS (frames per second) to 16.3 FPS, with using wavelet ``haar''.
}

\subsection{3D object classification}
\begin{figure}[bpt]
\color{black}
	\centering
	\subfigure[3D point cloud data.]
	{\includegraphics*[scale=0.9, viewport=60 500 165 623]{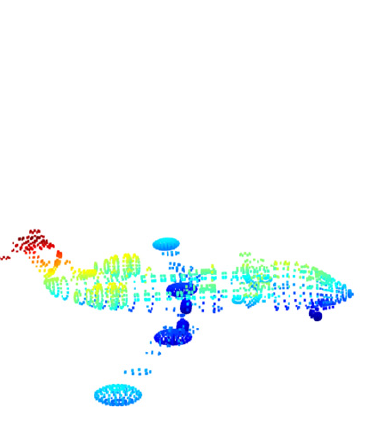}
	\label{fig_modelnet40_point_data}}\hspace{0pt}
	\subfigure[3D tensor.]
	{\includegraphics*[scale=1.05, viewport=225 500 360 623]{figures/modelnet40_example.pdf}
	\label{fig_modelnet40_tensor}}\hspace{0pt}
	\caption{\color{black}
    An example of 3D point cloud in ModelNet40.
    The 3D point cloud data is transformed to 3D tensor with size of $32\times32\times32$, as the input of deep networks.
    }
	\label{fig_modelnet40_example}
\end{figure}
\begin{table*}[!tbp]\color{black}
	%\scriptsize
    \centering
	%\small
	\caption{\color{black}Classification results on ModelNet40 of 3D WaveCNets.
    The symmetric Haar and Cohen wavelets can consistently improve the performance of 3D CNNs,
    while the improvements of asymmetric Daubechies wavelet are not guaranteed.}
	\label{Tab_3D_WaveCNet_accuracy}
		\setlength{\tabcolsep}{0.5mm}
			\begin{tabular}{c|ccccccccccc}\hline
\multirow{2}{*}{architecture}&3D CNN & \multicolumn{10}{c}{3D WaveCNet}\\\cline{3-12}
&(baseline)&haar& ch2.2& ch3.3& ch4.4& ch5.5& db2& db3& db4& db5& db6 \\\hline
3D VGG9     &$87.08$	&$88.01_{ +0.93}$	&$87.73_{+0.65}$	&$87.85_{+0.77}$	&$88.09_{+1.01}$	
&\textbf{88.25$_{+1.17}$}	&$87.40_{+0.32}$	&$87.36_{+0.28}$	&$86.06_{-1.02}$	&$84.97_{-2.11}$	&$81.43_{-5.65}$\\
3D ResNet16 &$84.32$	&$85.41_{+1.09}$	&\textbf{85.90$_{+1.58}$}	&$85.37_{+1.05}$	&$85.62_{+1.30}$	
&$85.86_{+1.54}$	&$85.62_{+1.30}$	&$85.70_{+1.38}$	&$85.43_{+1.11}$	&$85.33_{+1.01}$	&$85.13_{+0.81}$\\\hline
			\end{tabular}
\end{table*}
{\color{black}
We conduct 3D object classification experiments on 3D point cloud dataset ModelNet40 \cite{wu20153d}, using 3D VGG and ResNet.
ModelNet40 consists of 12311 (9843 for training and 2468 for test) 3D point cloud data captured from 40 categories of objects.
As shown in Fig. \ref{fig_modelnet40_point_data}, the point cloud data representing a 3D object contains a series of 3D coordinates.
As shown in Fig. \ref{fig_modelnet40_tensor}, we transform the 3D point cloud data to 3D tensor with size of $32\times32\times32$, as the input of deep networks.
We perform the classification experiments using 3D VGG9, 3D ResNet16, and their wavelet integrated versions (3D WVGG9 and WResNet16).
As 3D versions of VGG and ResNet, 3D VGG9 and ResNet16 are designed by removing the last two convolutional layers of 3D VGG11 and ResNet18.
The two 3D WaveCNets, 3D WVGG9 and WResNet16, apply 3D DWT to down-sample 3D feature maps.

Using the ModelNet40 training set, we train 3D WVGG9 and WResNet16 from scratch for 80 epochs, with batch size of 256 and initial learning rate of $0.1$.
The learning rate is dropped by a factor of 0.1 every 25 epochs.
For comparison, we also train the 3D VGG9 and 3D ResNet16, using the same hyper-parameter setting.
We evaluate their performance on the test set of ModelNet40, and list the results in Table \ref{Tab_3D_WaveCNet_accuracy}.
In Table \ref{Tab_3D_WaveCNet_accuracy}, ``3D CNN (baseline)'' corresponds to the results of original 3D VGG9 and ResNet16,
and ``3D WaveCNet'' represents that of 3D WVGG9 and WResNet16,
with different wavelets like Haar (``haar''), Cohen (``ch$p.\tilde{p}$''), and Daubechies (``db$p$'').
On the test set, the accuracy of 3D VGG9 and ResNet16 are 87.08\% and 84.32\%, respectively.
The integration of Cohen wavelet ``ch5.5'' and ``ch2.2'' improve the accuracy to 88.25\% and 85.90\%, respectively.
Similarly, Haar and Cohen wavelets also improve the performance of both 3D VGG9 and 3D ResNet16.
While all of different Daubechies wavelets could improve the accuracy of 3D ResNet16, ``db4'', ``db5'', and ``db6'' decrease the performance of 3D VGG9.

In summary, similar to the results of 2D image classification,
the symmetric Haar and Cohen wavelets can consistently improve the performance of 3D CNNs,
while the improvements of asymmetric Daubechies wavelet are not guaranteed.
}

\section{Conclusions}
In deep networks, the commonly used down-sampling operations result in the aliasing effects on the feature maps,
accumulating noise and breaking the object structures,
which leads to the weak noise-robustness of CNNs for image classification.
To suppress the aliasing effects, we design the general discrete wavelet transform (DWT) and inverse DWT (IDWT) layers,
and propose wavelet integrated convolutional networks (WaveCNets) by replacing the down-sampling operations in the common CNNs with DWT.
During the inference of WaveCNets, wavelet helps the networks to keep the basic object structures and resist the noise propagation.
WaveCNets achieve higher image classification accuracy,
better noise-robustness, and increased adversarial robustness
with various commonly used CNN architectures on ImageNet.
{\color{black}Our method can also improve the detection performance of faster R-CNN and RetinaNet on COCO benchmark.}

In future, we will exhaustively study the application of DWT/IDWT layers
in the image-to-image tasks.
We will also explore the applications of 3D DWT/IDWT integrated deep networks for 3D medical image processing.

%In future, we will extend our method to more network architectures for various image tasks,
%such as encoder-decoder for image segmentation.
%More applications like robustness to adversarial examples will be explored as well.

% Can use something like this to put references on a page
% by themselves when using endfloat and the captionsoff option.
\ifCLASSOPTIONcaptionsoff
  \newpage
\fi

% trigger a \newpage just before the given reference
% number - used to balance the columns on the last page
% adjust value as needed - may need to be readjusted if
% the document is modified later
%\IEEEtriggeratref{8}
% The "triggered" command can be changed if desired:
%\IEEEtriggercmd{\enlargethispage{-5in}}

% references section

% can use a bibliography generated by BibTeX as a .bbl file
% BibTeX documentation can be easily obtained at:
% http://mirror.ctan.org/biblio/bibtex/contrib/doc/
% The IEEEtran BibTeX style support page is at:
% http://www.michaelshell.org/tex/ieeetran/bibtex/
%\bibliographystyle{IEEEtran}
% argument is your BibTeX string definitions and bibliography database(s)
%\bibliography{IEEEabrv,../bib/paper}
%
% <OR> manually copy in the resultant .bbl file
% set second argument of \begin to the number of references
% (used to reserve space for the reference number labels box)
%\begin{thebibliography}{1}

%\bibitem{IEEEhowto:kopka}
%H.~Kopka and P.~W. Daly, \emph{A Guide to \LaTeX}, 3rd~ed.\hskip 1em plus
%  0.5em minus 0.4em\relax Harlow, England: Addison-Wesley, 1999.

%\end{thebibliography}

{\small
\bibliographystyle{IEEEtran}
\bibliography{IEEEtran}
}

% biography section
%
% If you have an EPS/PDF photo (graphicx package needed) extra braces are
% needed around the contents of the optional argument to biography to prevent
% the LaTeX parser from getting confused when it sees the complicated
% \includegraphics command within an optional argument. (You could create
% your own custom macro containing the \includegraphics command to make things
% simpler here.)
%\begin{IEEEbiography}[{\includegraphics[width=1in,height=1.25in,clip,keepaspectratio]{mshell}}]{Michael Shell}
% or if you just want to reserve a space for a photo:

\begin{IEEEbiography}{Qiufu Li}
received the B.S. degree in mathematics and information science from North Minzu University, Yinchuan, China,
M.S. degree in wavelet analysis and signal processing from North Minzu University,
and Ph.D degree from Beijing Insitute of Technology, Beijing China, in 2010, 2013, and 2017, respectively.
He has been a postdoctoral fellow at School of Computer Science and Software Engineering, Shenzhen University, Shenzhen, China.
He is currently research associate fellow at the same school.
His research interests are in the areas of wavelet analysis, digital image processing, and deep learning.
\end{IEEEbiography}

% if you will not have a photo at all:
\begin{IEEEbiography}{Linlin Shen}
is currently the ``Pengcheng Scholar'' Distinguished Professor at School of Computer Science and Software Engineering, Shenzhen University, Shenzhen, China.
He is also an Honorary professor at School of Computer Science, University of Nottingham, UK.
He serves as the director of Computer Vision Institute, AI Research Center for Medical Image Analysis \& Diagnosis and China-UK joint research lab for visual information processing.
He also serves as the Co-Editor-in-Chief of the IET journal of Cognitive Computation and Systems.
He received the BSc and MEng degrees from Shanghai JiaoTong University,
Shanghai, China, and the Ph.D. degree from the University of Nottingham, Nottingham, U.K.
He was a Research Fellow with the University of Nottingham, working on MRI brain image processing.
His research interests include deep learning, facial analysis and medical image processing. Prof.
Shen is listed as the Most Cited Chinese Researcher by Elsevier.
He received the Most Cited Paper Award from the journal of Image and Vision Computing.
His cell classification algorithms were the winners of the International Contest on Pattern Recognition Techniques for Indirect Immunofluorescence Images held by ICIP 2013 and ICPR 2016.
\end{IEEEbiography}

% insert where needed to balance the two columns on the last page with
% biographies
%\newpage

\begin{IEEEbiography}{Sheng Guo}
received the Ph.D degree from Shenzhen Institutes of Advanced Technology, Chinese Academy of Sciences, Shenzhen, China, in 2017.
He is currently a senior algorithm expert with Ant Financial.
His current research interests are object classification, scene/object detection, video classification.
He was the first runner-up at the ImageNet Large Scale Visual Recognition Challenge 2015 in scene recognition and winner at WebVision Challenge 2017.
\end{IEEEbiography}

\begin{IEEEbiography}{Zhihui Lai}
received the B.S. degree in mathematics from South China Normal University, M.S. degree from Jinan University,
and the Ph.D degree in pattern recognition and intelligence system from Nanjing University of Science and Technology (NUST), China, in 2002, 2007 and 2011, respectively.
He has been a research associate, postdoctoral fellow and research fellow since 2010 at The Hong Kong Polytechnic University.
He also has been a Postdoctoral Fellow at Bio-Computing Research Center, Shenzhen Graduate School, Harbin Institute of Technology (HIT) for 2011-2013.
He has published over 100 scientific articles, including more than 30 IEEE transactions.
He won the best paper award in 2008 Chinese Conference on Pattern Recognition. He was recognized as Overseas High-Caliber Personnel of Shenzhen (Peacock Plan) in 2015.
Now, he is a professor of the College of Computer Science and Software Engineering, Shenzhen University.
Dr. Lai is a reviewer of more than 20 international journals, including the IEEE TPAMI, TNNLS, TIP, TCSVT, T-Cybernetics, Pattern Recognition, Information Sciences and so on.
He is currently an Associate Editor of the International Journal of Machine Learning and Cybernetics.
\end{IEEEbiography}

% You can push biographies down or up by placing
% a \vfill before or after them. The appropriate
% use of \vfill depends on what kind of text is
% on the last page and whether or not the columns
% are being equalized.

%\vfill

% Can be used to pull up biographies so that the bottom of the last one
% is flush with the other column.
%\enlargethispage{-5in}

% that's all folks
\end{document}